\pdfoutput=1

\documentclass[11pt]{article}

\usepackage{acl}

\usepackage[T1]{fontenc}

\usepackage[utf8]{inputenc}
\usepackage{graphics} 

\usepackage{microtype}
\usepackage{times}
\usepackage{latexsym}
\newcommand{\result}[2]{ #1 \color{gray}{\scriptsize{$#2$}}}
\usepackage{algorithm}
\usepackage{algorithmicx}
\usepackage{algpseudocode}
\usepackage{amssymb}
\usepackage{makecell}
\usepackage{graphicx}
\usepackage{amsmath}
\usepackage{mathrsfs}
\usepackage{amsthm}
\usepackage{natbib}  
\usepackage{caption} 
\usepackage{changes}
\newtheorem*{remark}{Remark}
\usepackage{hyperref}
\usepackage{subcaption}
\usepackage{multirow}
\usepackage{footnote}

\usepackage{tikzsymbols}

\title{Learning Disentangled Textual Representations via \\ Statistical Measures of Similarity}

\author{\textbf{Pierre Colombo\textsuperscript{\rm 1}, Guillaume  Staerman\textsuperscript{\rm 2}}, \textbf{Nathan Noiry\textsuperscript{\rm 2 3},  Pablo Piantanida\textsuperscript{\rm 4}} \\
\textsuperscript{\rm 1}L2S,CentraleSupelec CNRS Universite Paris-Saclay,\\
\textsuperscript{\rm 2} LTCI, Telecom Paris, Institut Polytechnique de Paris, \\
\textsuperscript{\rm 3} althiqa, \\
\textsuperscript{\rm 4} ILLS, Université McGill - ETS - MILA - CNRS - Université Paris-Saclay - CentraleSupélec \\
\textsuperscript{\rm 1}pierre.colombo@centralesupelec.fr \\
}

\begin{document}
\maketitle

\begin{abstract}
When working with textual data, a natural application of disentangled representations is fair classification where the goal is to make predictions without being biased (or influenced) by sensitive attributes that may be present in the data (e.g., age, gender or race). Dominant approaches to disentangle a sensitive attribute from textual representations rely on learning simultaneously a penalization term that involves either an adversarial loss (e.g., a discriminator) or an information measure (e.g., mutual information). However, these methods require the training of a deep neural network with several parameter updates for each update of the representation model. As a matter of fact, the resulting nested optimization loop is both time consuming, adding complexity to the optimization dynamic, and requires a fine hyperparameter selection (e.g., learning rates, architecture). In this work, we introduce a family of regularizers for learning disentangled representations that do not require training.  These regularizers are based on statistical measures of similarity between the conditional probability distributions with respect to the sensitive attributes. Our novel regularizers do not require additional training, are faster and do not involve additional tuning while achieving better results both when combined with pretrained and randomly initialized text encoders.
\end{abstract}

\section{Introduction}\label{sec:introduction}
\begin{figure}%
    \centering
    \subfloat[]{{\includegraphics[width=2.6cm]{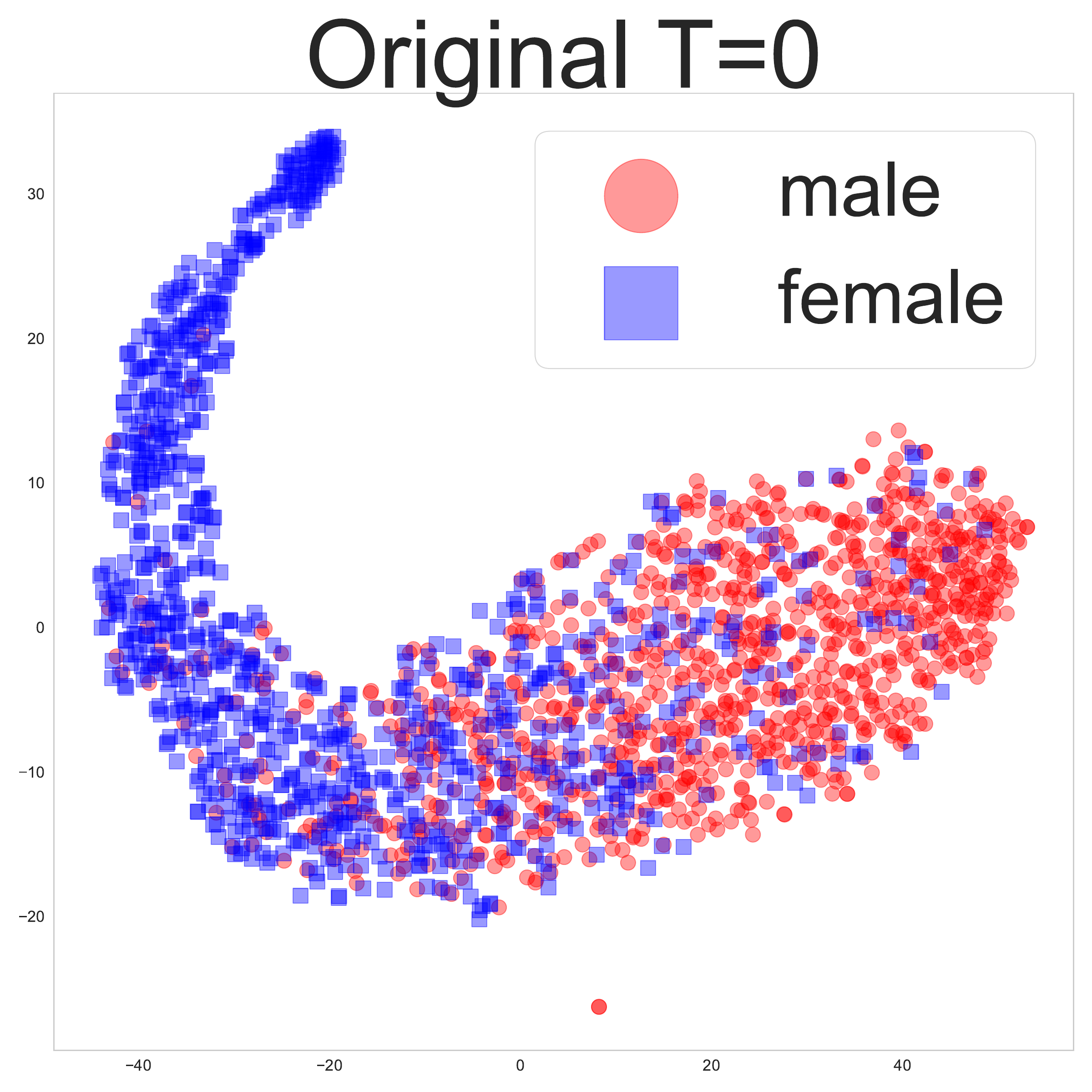} }}%
    \subfloat[]{{\includegraphics[width=2.6cm]{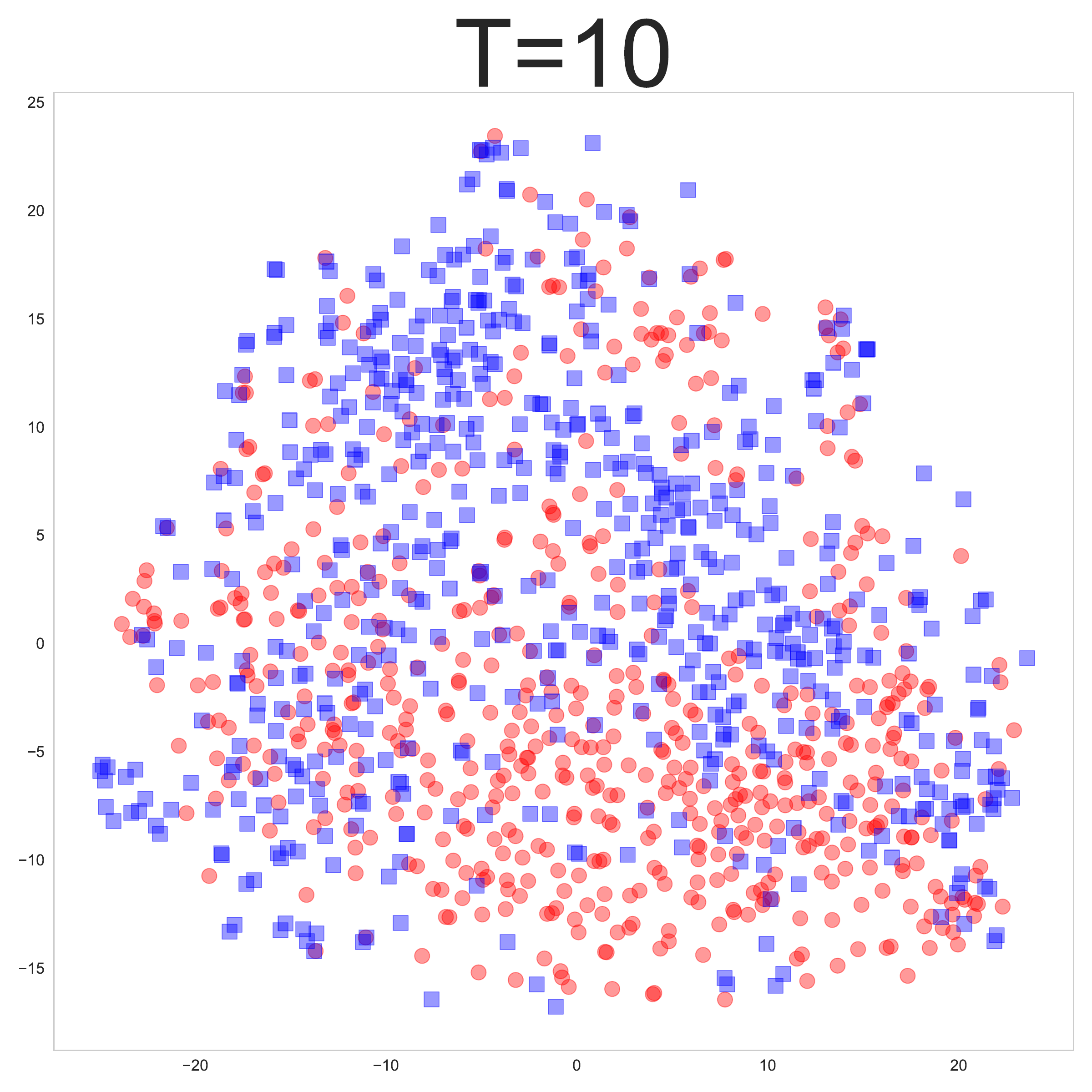} }}
    \subfloat[]{{\includegraphics[width=2.6cm]{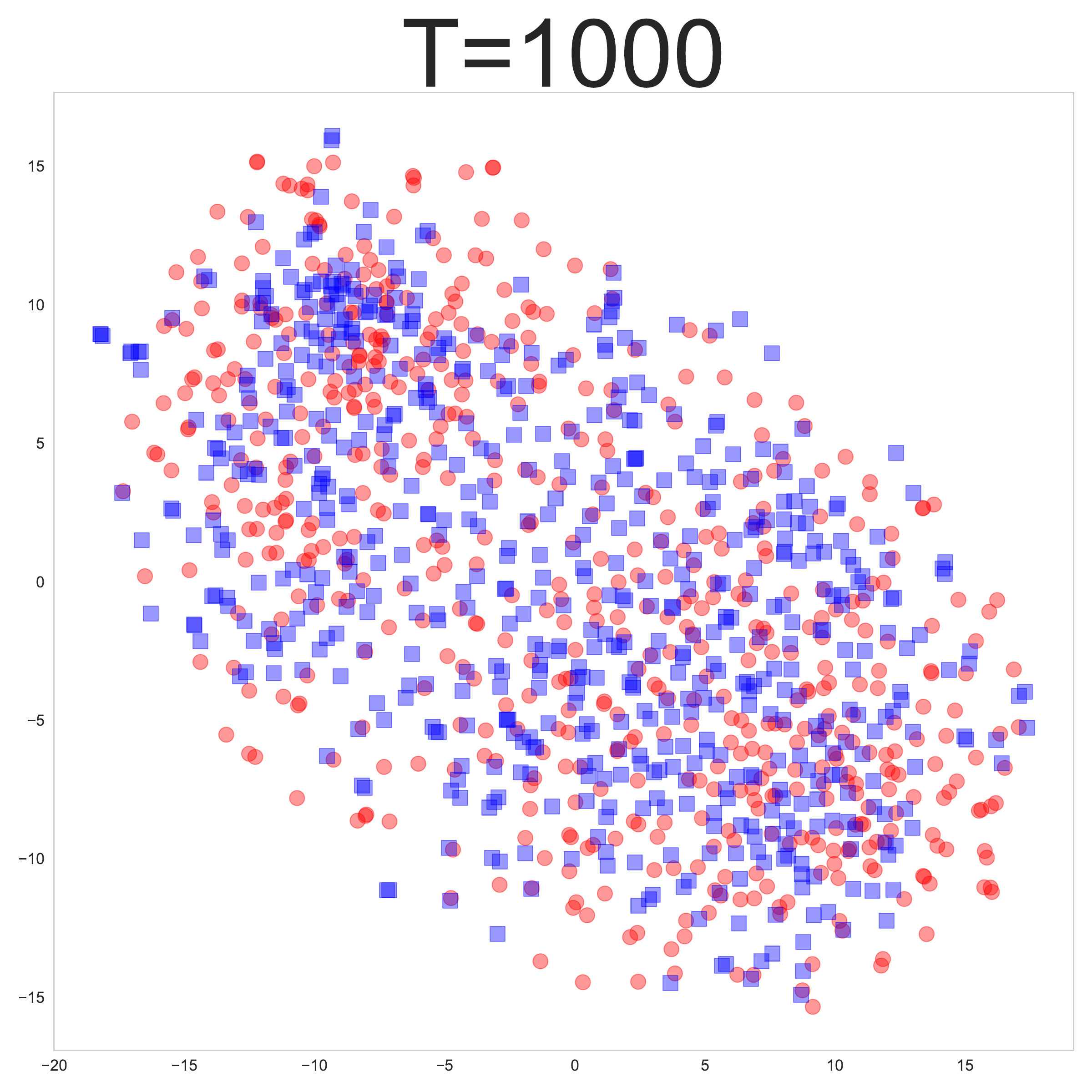} }}%
    \caption{PCA followed by a T-SNE projection of BERT embeddings of the sentences of \texttt{DIAL} corpus after T=0,10,1000 iterations of our framework (based on Sinkhorn divergence). Colors display the sensitive (i.e., binary gender) attribute.}   \vspace{-.5cm}
    \label{fig:exampleDisentangledRepresentation}
\end{figure}

    As natural language processing (NLP) systems are taken up in an ever wider array of sectors (e.g., legal system \cite{dale2019law}, insurance \cite{ly2020survey}, education \cite{litman2016natural}, healthcare \cite{basyal2020systematic}), there are growing concerns about the harmful potential of \textit{bias} in such systems \cite{leidner2017ethical}. Recently, a large body of research aims at analyzing, understanding and addressing \textit{bias} in various applications of NLP including language modelling \cite{liang2021towards}, machine translation \cite{stanovsky2019evaluating}, toxicity detection \cite{dixon2018measuring} and classification \cite{adversarial_removal}. In NLP, current systems often rely on learning continuous embedding of the input text. Thus, it is crucial to ensure that the learnt continuous representations do not exhibit \textit{bias} that could cause representational harms \cite{blodgett2020language,barocas2017problem}, i.e.,  representations less favourable to specific social groups. One way to prevent the aforementioned phenomenon is to enforce disentangled representations, i.e., representations that are independent of a sensitive attribute (see Fig. ~\ref{fig:exampleDisentangledRepresentation} for a visualization of different degrees of disentangled representations).

\noindent Learning disentangled representations has received a growing interest as it has been shown to be useful for a wide variety of tasks (e.g., style transfer \cite{style_transfert_1}, few shot learning \cite{karn2021few}, fair classification \cite{colombo2021novel}). For text, the dominant approaches to learn such representations can be divided into two classes. The first one, relies on an adversary that is trained to recover the discrete sensitive attribute from the latent representation of the input \cite{adv_classif_fair_1}. However, as pointed out by \citet{adversarial_removal_2}, even though the adversary seems to do a perfect job during training, a fair amount of the sensitive information can be recovered from the latent representation when training a new adversary from scratch. The second line of research involves a regularizer that is a trainable surrogate of the mutual information (MI) (e.g., CLUB \cite{cheng2020club}, MIReny \cite{colombo2021novel}, KNIFE \cite{pichler2020estimation}, MINE \cite{mine,DBLP:conf/emnlp/ColomboCLC21}) and achieves higher degrees of disentanglement. However, as highlighted by recent works \cite{mcallester2020formal,song2019understanding}, these estimators are hard to use in practice and the optimization procedure (see App.~\ref{subsec:oldalgo}) involves several updates of the regularizer parameters at each update of the representation model. As a consequence, these procedures are both time consuming and involve extra hyperparameters (e.g., optimizer learning rates, architecture, number of updates of the nested loop) that need to be carefully selected which is often not such an easy task.

\noindent \textbf{Contributions.} In this work, we focus our attention on learning to disentangle textual representations from a discrete attribute. Our method relies on a novel family of regularizers based on discrepancy measures. We evaluate both the disentanglement and representation quality on fair text classification. Formally, our contribution is two-fold:\\
\indent (1) \textit{A novel formulation of the problem of learning disentangled representations.} Different from previous works--either minimizing a surrogate of MI or training an adversary--we propose to  minimize a statistical measure of similarity between the underlying probability distributions conditioned to the sensitive attributes. This novel formulation allows us to derive new regularizers with convenient   properties:  (i) not requiring additional learnable parameters; (ii)  alleviating  computation burden; and (iii) simplifying the optimization dynamic.
\\ \indent (2) \textit{Applications and numerical results.} We carefully evaluate our new framework on four different settings coming from two different datasets. We strengthen the experimental protocol of previous works \cite{colombo2021novel,ravfogel2020null} and test our approach both on randomly initialized encoder (using RNN-based encoder) and during fine-tuning of deep contextualized pretrained representations\footnote{Previous works (e.g., \cite{ravfogel2020null}) do not fine-tune the pretrained encoder when testing their methods.}. Our experiments are conducted on four different main/sensitive attribute pairs and involve the training of over $280$ deep neural networks. Our findings show that: (i) disentanglement methods behave differently when applied to randomly initialized or to deep contextualized pretrained encoder; and (ii) our framework offers a better accuracy/disentanglement trade-off than existing methods (i.e., relying on an adversary or on a MI estimator) while being faster and easier to train. Model, data and code are available at \url{https://github.com/PierreColombo/TORNADO}.




\section{Related Work}\label{sec:rw}

Considering a tuple $(X,S)$ where $X$ is a random variable (r.v.) defined on the space of text $\mathcal{X}$ and $S$ is a binary r.v. which corresponds to a sensitive attribute. Learning disentangled representations aims at learning the parameter $\theta$ of the encoder $f_\theta : \mathcal{X} \rightarrow \mathcal{Z} \subset \mathbb{R}^d$ which maps  $X$ to a latent representation $Z = f_\theta(X) \in \mathbb{R}^d$, where $d \in \mathbb{N}_*$ corresponds to the dimension of the embedding space. The goal is that $Z$ retains as much useful information from $X$ while being oblivious of $S$. Among the numerous possible applications for disentangled representations, we choose to focus on fair classification as it is a natural task to define the aforementioned useful information. In the fair classification task, we assume access to $Y$, a binary r.v., which corresponds to the main label/attribute. In order to learn disentangled representations for fair classification, we follow previous works \cite{adv_classif_fair,cheng2020improving} and we will be minimizing the loss $\mathcal{L}(\phi,\psi,\theta)$, which is defined as follows:

\vspace{-0.5cm}
\begin{equation}\label{eq:all_loss}
\underbrace{\textrm{CE}\big(C_\phi(f_\theta(X)),Y\big)}_{\textrm{target task}} + \lambda \cdot \underbrace{R\big(f_\theta(X),S;\psi\big)}_{\textrm{regularizer}}, 
\end{equation} 

\noindent where $C_\phi : \mathcal{Z} \rightarrow \mathcal{Y}$ refers to the main classifier; $\phi$ to its learnable parameters; CE to the cross-entropy loss; R denotes the disentanglement regularizer; $\psi$ its parameters and $\lambda$ controls the trade-off between disentanglement and success in the classification task. We next review the two main methods that currently exist for learning textual disentangled representations: \emph{adversarial-based} and \emph{MI-based}.




\subsection{Adversarial-Based Regularizers}
In the context of disentangled representation learning, a popular method is to rely on adding an adversary to the encoder (e.g., texts \cite{coavoux2018privacy}, images \cite{adv_classif_fair_1}, categorical data \cite{adv_classif_fair}). This adversary is competing  against the encoder trying to learn the main task objective. In this line of work, $R(f_\theta(X),S;\psi) = - \textrm{CE}(C_\psi(f_\theta(X)),S)$ where $C_\psi : \mathcal{Z} \rightarrow \mathcal{S}$ refers to the adversarial classifier that is trained to minimize $\textrm{CE}(C_\psi(f_\theta(X)),S)$. Denoting by $\mathbb{P}_{Z|S=0}$ and $\mathbb{P}_{Z|S=1}$ the probability distribution of the conditional r.v. $Z|S=0$ and $Z|S=1$, respectively, these works build on the fact that if $\mathbb{P}_{Z|S=0}$ and $\mathbb{P}_{Z|S=1}$ are different, the optimal adversary will be able to recover sensitive information from the latent code $Z$. Although adversaries have achieved impressive results in many applications when applied to attribute removal, still a fair amount of information may remain in the latent representation \cite{multiple}. 
\subsection{MI-Based Regularizers}
To better protect sensitive information, the second class of methods involves direct mutual information  minimization. MI lies at the heart of information theory and measures statistical dependencies between two random variables $Z$ and $S$ and find many applications in machine learning \cite{boudiaf2020information,boudiaf2020unifying,boudiaf2021mutual}. The MI is a non-negative quantity that is 0 if and only if $Z$ and $S$ are independent and is defined as follows:
    \begin{align}\label{eq:mi_def}
    I(Z;S) 
     & = \text{KL}(\mathbb{P}_{ZS} \| \mathbb{P}_{Z}  \otimes \mathbb{P}_S) ,
     \end{align}

\noindent where the joint probability distribution of $(Z,S)$ is denoted by $\mathbb{P}_{ZS}$; marginals of $Z$ and $S$ are denoted by $\mathbb{P}_Z$ and $\mathbb{P}_S$ respectively; and KL stands for the Kullback-Leibler divergence.  Although computing the MI is challenging \cite{estimation,pichler2022knife}, a plethora of recent works devise new lower \cite{mine,nce} and upper bounds \cite{cheng2020club,colombo2021novel} ${\Tilde{I}_\psi(f_\theta(X);S)}$ where $\psi$ denotes the trainable parameters of the surrogate of the MI. In that case, $R(f_\theta(X),S;\psi) = {\Tilde{I}_\psi(f_\theta(X);S)}$. These methods build on the observation that if  $I(Z;X) >  0$ then  $\mathbb{P}_{Z|S=0}
$ and $\mathbb{P}_{Z|S=1}$ are different and information about the sensitive label $S$ remains in $Z$. Interestingly, these approaches achieve better results than adversarial training on various NLP tasks \cite{cheng2020improving} but involve the use of additional (auxiliary) neural networks. 

\subsection{Limitations of Existing Methods} The aforementioned methods involve the use of extra parameters (i.e., $\psi$) in the regularizer. As the regularizer computes a quantity based on the representation given by the encoder with parameter $\theta$, any modification of $\theta$ requires an adaptation of the parameter of $R$ (i.e., $\psi$). In practice, this adaptation is performed using gradient descent-based algorithms and requires several gradient updates. Thus, a nested loop (see App. \ref{subsec:oldalgo}) is needed. Additional optimization parameters and the nested loop both induce additional complexity and require a fine-tuning which makes these procedures hard to be used on large-scale datasets. To alleviate these issues, the next section describes a parameter-free framework to get rid of the $\psi$ parameter present in $R$.

\section{Proposed Method}\label{sec:fc}
This section describes our approach to learn disentangled representations. We first introduce the main idea and provide an algorithm to implement the general loss. We next describe the four similarity measures proposed in this approach.

\subsection{Method Overview}

As detailed in Section \ref{sec:rw}, existent methods generally rely on the use of neural networks either in the form of an adversarial regularizer or to compute upper/lower bounds of the MI between the embedding $Z=f_{\theta}(X)$ and the sensitive attribute $S$. Motivated by reducing the computational and complexity load, we aim at providing regularizers that are light and easy to tune. To this end, we need to get rid of the nested optimization loop, which is both time consuming and hard to tune in practice
 since the regularizer contains a large number of parameters (e.g., neural networks) that need to be trained by gradient descent. Contrarily to previous works in the literature, and following the intuitive idea that $\mathbb{P}_{Z|S=0}$ and $\mathbb{P}_{Z|S=1}$ should be as close as possible, we introduce similarity measures between $\mathbb{P}_{Z|S=0}$ and $\mathbb{P}_{Z|S=1}$ to build a regularizer $R$.  It is worth noting that the similarity measures do not require any additional learnable parameters. For the sake of clarity, in the reminder of the paper we define $\mathbb{P}_{i} \triangleq  \mathbb{P}_{Z|S=i}$ and $Z_i\triangleq f_{\theta}(X|S=i)$ for $i\in \{0,1 \}$. Given a similarity measure defined as  $\text{SM}: \mathcal{M}_{+}^1(\mathcal{Z}) \times \mathcal{M}_{+}^1 (\mathcal{Z})  \longrightarrow \mathbb{R}_+ $ where $\mathcal{M}_{+}^1 (\mathcal{Z})$ denotes the space of probability distributions on $\mathcal{Z}$,  we propose to regularize the downstream task by  $\text{SM}(\mathbb{P}_0, \mathbb{P}_1)$. Precisely, the optimization problem boils down to the following objective: 
\begin{equation}\label{eq:problem}
    \mathcal{L}(\phi,\theta) =  \underbrace{\textrm{CE}(C_\phi(f_\theta(X)),Y)}_{\textrm{target task}} + \lambda \cdot \underbrace{\text{SM}(\mathbb{P}_{0}, \mathbb{P}_{1})}_{\textrm{regularizer}}. 
\end{equation}

The proposed statistical measures of similarity, detailed in Section \ref{ssec:similarity_function}, have explicit and simple formulas. It follows that the use of neural networks is no longer necessary in the regularizer term which reduces drastically  the complexity of the resulting learning problem. The disentanglement can be controlled by selecting appropriately the measure SM. For the sake of place, the algorithm we propose to solve (\ref{eq:problem}) is deferred to the App.~\ref{sec:algo}.

\subsection{Measure of Similarity between Distributions}\label{ssec:similarity_function}

In this work, we choose to focus on four different (dis-) similarity functions ranging from the most popular in machine learning such as the Maximum Mean Discrepancy measure ({MMD}) and the Sinkhorn divergence ({SD}) to standard statistical discrepancies such as the Jeffrey  divergence ({J}) and  the Fisher-Rao distance ({FR}).


\subsubsection{Maximum Mean Discrepancy.}
Let $k:\mathcal{Z}\times \mathcal{Z}\rightarrow \mathbb{R}$ be a kernel and $\mathcal{H}$ its corresponding Reproducing Kernel Hilbert Space with inner product $\langle ., . \rangle_{\mathcal{H}}$ and norm $\|.\|_{\mathcal{H}}$. Denote by $\mathcal{B}_{\mathcal{H}}=\{f\; | \; \|f\|_{\mathcal{H}} \leq 1 \}$ the unit ball of $\mathcal{H}$. The Maximum Mean Discrepancy ({MMD}) \cite{gretton} between the two conditional distributions $\mathbb{P}_0, \mathbb{P}_1 \in \mathcal{M}^1_+(\mathcal{Z})$  associated with the kernel $k$, is defined as: 
\begin{align*}
    \text{MMD}(\mathbb{P}_0,\mathbb{P}_1) &= \hspace{-0.1cm}\underset{\Psi \in \mathcal{B}_{\mathcal{H}}}{\sup} \big \lvert \mathbb{E}_{\mathbb{P}_0}[\Psi(Z_0)] \hspace{-0.05cm} - \hspace{-0.05cm}\mathbb{E}_{\mathbb{P}_1}[\Psi(Z_1)] \big \rvert \\&=
    \mathbb{E}_{\mathbb{P}_0 \otimes \mathbb{P}_0} [k(Z_0,Z_0')] \\& \quad +\mathbb{E}_{\mathbb{P}_1 \otimes \mathbb{P}_1} [k(Z_1,Z_1')] \\& \quad-2\,\mathbb{E}_{\mathbb{P}_0 \otimes \mathbb{P}_1} [k(Z_0,Z_1)].
\end{align*}

\noindent The {MMD} can be estimated with a quadratic computational complexity $\mathcal{O}(n^2)$ where $n$ is the sample size. In this paper, {MMD} is computed using the Gaussian kernel $k:(z_0,z_1)\mapsto \exp(-\|z_0-z_1\|^2 / 2\sigma^2)$, where $\|\cdot\|$ is the usual euclidean norm.

\subsubsection{Sinkhorn Divergence.}
The Wasserstein distance aims at comparing two probability distributions through the resolution of the Monge-Kantorovich mass transportation problem (see e.g. ~\citet{villani,peyre}):
\begin{equation}\label{OT-primal}
\hspace{-0.1em}\text{W}^p(\mathbb{P}_0, \mathbb{P}_1) \hspace{-0.1em}= \hspace{-1em}\underset{ \pi~\in~\mathcal{U}(\mathbb{P}_0, \mathbb{P}_1)}{\min} \hspace{-0.2em} \int_{\mathcal{Z} \times \mathcal{Z}}  \hspace{-0.5cm} \|z_0-z_1\|^p d\pi(z_0, z_1),
\end{equation}
where $\mathcal{U}(\mathbb{P}_0, \mathbb{P}_1)= \{ \pi \in \mathcal{M}^1_+(\mathcal{Z} \times \mathcal{Z}): \; \; \int \pi(z_0,z_1)dy =\mathbb{P}_0(z_0) ;\int \pi(z_0,z_1) dx=\mathbb{P}_1(z_1) \}$  is the set of joint probability distributions with marginals $\mathbb{P}_0$ and $\mathbb{P}_1$. For the sake of clarity, the power $p$ in W$^p$ is omitted in the remainder of the paper. When $\mathbb{P}_0$ and $\mathbb{P}_1$ are discrete measures, (\ref{OT-primal}) is a linear problem and can be solved with a supercubic complexity $\mathcal{O}(n^3\log(n))$, where $n$ denotes the sample size. 
To overcome this computational drawback, \citet{cuturi13} added an entropic regularization term to the transport cost to obtain a strongly convex problem  solvable using the Sinkhorn-Knopp  algorithm \cite{sinkhorn1964} leading to a computational cost of $\mathcal{O}(n^2)$. The bias introduced by the regularization term, i.e., the quantity is not longer zero when comparing to the same probability distribution, have been  corrected by \citet{genevay18b} leading to the known Sinkhorn Divergence ({SD}) defined as: 
\begin{equation*}
    \text{{SD}}_{\varepsilon} (\mathbb{P}_0,\mathbb{P}_1)= \text{W}_{\varepsilon}(\mathbb{P}_0,\mathbb{P}_1) - \frac{1}{2} \sum_{i=0}^{1} \text{W}_{\varepsilon}(\mathbb{P}_i,\mathbb{P}_i),
\end{equation*}
\noindent where  $\text{W}_{\varepsilon}(\mathbb{P}_0,\mathbb{P}_1)$ is equal to
\begin{align*}
& \underset{ \pi~\in~\mathcal{U}(\mathbb{P}_0,\mathbb{P}_1)}{\min} \hspace{-0.2em}  \int_{\mathcal{Z} \times \mathcal{Z}}   \|z_0-z_1\|^p d\pi(z_0, z_1)+\varepsilon \, \text{H}(\pi), 
\end{align*}
\noindent with $\text{H}(\pi)=\int \pi(z_0,z_1) \log(\pi(z_0,z_1))dz_0dz_1$.

\subsubsection{Fisher-Rao Distance.} The Fisher-Rao distance ({FR}) \citep{Rao1945} is a Riemannian metric defined on the space of parametric distributions relying on the Fisher information. The Fisher information matrix provides a natural Riemannian structure \cite{amari2012differential}. It is known to be more accurate than popular divergence measures \cite{costa2015fisher}. Let $\mathcal{M}_+^1(\mathcal{Z}, \mathcal{P})$ be the family of parametric distributions with the parameter space $\mathcal{P} \subset \mathbb{R}^d$. The {FR} distance is defined as the geodesic distance \footnote{The geodesic is the curve that provides the shortest length.} between elements (i.e., probability measures) on the manifold $\mathcal{M}_+^1(\mathcal{Z}, \mathcal{P})$.  Parametrizing $\mathbb{P}_0,\mathbb{P}_1$ by parameters $p_1,p_2 \in \mathcal{P}$,  respectively, such that $\mathbb{P}^{p_0}_0\triangleq \mathbb{P}_0 $ and $\mathbb{P}^{p_1}_1\triangleq \mathbb{P}_1$, the {FR} distance between  $\mathbb{P}_0^{p_0}$ and $\mathbb{P}_1^{p_1}$ is defined as:
\begin{equation}\label{eq:FR}
\text{FR}(\mathbb{P}_0^{p_0},\mathbb{P}_1^{p_1}) = \min_{\gamma} \int | \sqrt{\gamma^\prime(t)^\top G(p_0,p_1) \gamma(t)} | dt
\end{equation} where $\gamma(t)$ is the curve connecting $p_0$ and $p_1$ in the parameter space $\mathcal{P}$; and $G(p_0,p_1)$ is the Fisher information matrix of $(p_0,p_1)$. In general, the optimization problem of (\ref{eq:FR}) can be solved using the well-known Euler-Lagrange differential equations leading to computational difficulties. \citet{Atkinson} have provided computable closed-form for specific families of distributions such as Multivariate Gaussian with diagonal covariance matrix. Under this assumption, the parameters $p_0$ and $p_1$ are defined by $p_{i,j}=(\mu_{i,j},\sigma_{i,j})\in \mathbb{R}^2$ for $i\in \{0,1\}$ and $1\leq j\leq d$ with $\mu_i\in \mathbb{R}^d$ the mean vector and $\text{Diag}(\sigma_{i})$ the diagonal covariance matrix of $\mathbb{P}_i$  where $\sigma_i$ is the variance vector. The resulting {FR} metric admits the following closed-form (see e.g. \citet{Pinele2020}:
\begin{align*}
    \text{FR}(\mathbb{P}_0^{p_0}, \mathbb{P}_1^{p_1})= \sqrt{\sum_{j=1}^{d} \left[d_{\text{FR}}(p_{0,j},p_{1,j})\right]^2},
\end{align*}
\noindent where $d_{\text{FR}}(p_{0,j},p_{1,j})$ is the univariate Fisher-Rao detailed in the App.~\ref{subsec:FR} for the sake of space.


\subsubsection{Jeffrey Divergence.}
The Jeffrey divergence ({J}) is a symmetric version of the Kullback-Leibler ({KL}) divergence and measures the similarity between two probability distributions. Formally, it is defined as follow: 
\begin{align*}
    \text{J}(\mathbb{P}_0,\mathbb{P}_1)= \frac{1}{2} & \big[ \text{KL}(\mathbb{P}_0 \| \mathbb{P}_1) +  \text{KL}(\mathbb{P}_1  \| \mathbb{P}_0 ) \big ].
\end{align*}

\noindent Computing the $\text{KL}(\mathbb{P}_0 \| \mathbb{P}_1)$ either requires to have knowledge of $\mathbb{P}_0$ and $\mathbb{P}_1$, or to have knowledge about the density ratio \cite{rubenstein2019practical}. Without any further assumption on  $\mathbb{P}_0$, $\mathbb{P}_1$ or the density ratio, the resulting inference problem is known to be provably hard \cite{nguyen2010estimating}. Although previous works have addressed the estimation problem without making assumptions on $\mathbb{P}_0$ and $\mathbb{P}_1$ \cite{nce,evo_mine,mine}, these methods often involve additional parameters (e.g., neural networks \cite{song2019understanding}, kernels \cite{mcallester2020formal}), require additional tuning \cite{hershey2007approximating}, and are time expensive. Motivated by speed, simplicity and to allow for fair comparison with {FR}, for this specific divergence, we choose to make the assumption that $\mathbb{P}_0$ and $\mathbb{P}_1$ are multivariate Gaussian distributions with mean vector $\mu_0$ and $\mu_1$ and diagonal covariance matrices: $\Sigma_0$ and $\Sigma_1$. Thus, $\text{KL}(\mathbb{P}_0,\mathbb{P}_1)$ boils down to:

\begin{align*}\label{eq:kl}
\log \hspace{-0.1cm}\frac{|\Sigma_0|}{|\Sigma_1|}\hspace{-.05cm}-\hspace{-.1cm}d\hspace{-.1cm}+\hspace{-.1cm}\text{Tr}(\Sigma_0^{-1}\hspace{-0.05cm}\Sigma_1) \hspace{-.1cm} + \hspace{-.1cm}(\mu_0\hspace{-.1cm}-\hspace{-.1cm}\mu_1)^{T} \Sigma_0^{-1}(\mu_0\hspace{-.1cm}-\hspace{-.1cm}\mu_1),
\end{align*}
where $\text{Tr}(\Sigma_0^{-1}\Sigma_1)$ is the trace of $\Sigma_0^{-1}\Sigma_1$.

\begin{remark}
FR and J are computed under the multivariate Gaussian with diagonal covariance matrix assumption. In this case, the Sinkhorn approximation is not needed as (\ref{OT-primal}) can be efficiently computed thanks to the following closed-form:
\begin{equation*}
    \mathrm{W}(\mathbb{P}_0,\hspace{-0.05cm} \mathbb{P}_1)\hspace{-.1cm}=\hspace{-.1cm}\|\mu_0 -\mu _1 \hspace{-0.05cm}\|^{2}\hspace{-.1cm}+ \hspace{-.1cm}\mathrm{Tr}\hspace{-.05cm}\left(\Sigma_0\hspace{-.1cm}+\hspace{-.1cm}\Sigma_1\hspace{-.1cm}-\hspace{-.1cm} 2(\Sigma_0 \Sigma_1 )^{1/2} \hspace{-0.05cm}\right)
\end{equation*}
\end{remark}

\begin{remark}
Quantities defined in this section are replaced by their empirical estimate. Due to space constraints, the formula are described in App.~\ref{subsec:emp}.
\end{remark}

\section{Experimental Setting}
In this section, we describe the datasets, metrics, encoder and baseline choices. Additional experimental details can be found in App.~\ref{sec:supp_experimental_details}. For fair comparison, all models were re-implemented.

\subsection{Datasets} To ensure backward comparison with previous works, we choose to rely on the \texttt{DIAL} \cite{dial} and  the \texttt{PAN}  \cite{rangel2014overview} datasets. For both, main task labels ($Y$) and sensitive labels ($S$) are binary, balanced and splits follow \cite{adversarial_removal_2}. Random guessing is expected to achieve near 50\% of accuracy. 
\\\noindent The \texttt{DIAL} corpus has been automatically built from tweets and the main task is either polarity\footnote{Polarity or emotion have been widely studied in the NLP community \cite{jalalzai2020heavy,colombo2019affect}} or mention prediction. The sensitive attribute is related to race (i.e., non-Hispanic blacks and non-Hispanic whites) which is obtained using the author geo-location and the words used in the tweet. 
\\\noindent The \texttt{PAN} corpus is also composed of tweets and the main task is to predict a mention label. The sensitive attribute is obtained through a manual process and annotations contain the age and gender information from 436 Twitter users.

\subsection{Metrics}
For the choice of the evaluation metrics, we follow the experimental setting of \citet{colombo2021novel,adversarial_removal,coavoux2018privacy}. To measure the success of the main task, we report the classification accuracy. To measure the degree of disentanglement of the latent representation we train from scratch an adversary to predict the sensitive labels from the latent representation. In this framework, a perfect model would achieve a high main task accuracy (i.e., near 100\%)  and a low (i.e., near 50\%) accuracy as given by the adversary prediction on the sensitive labels. Following \citet{colombo2021novel}, we also report the disentanglement dynamic following variations of  $\lambda$ and train a different model for each $\lambda \in [0.001,0.01,0.1,1,10]$.

\subsection{Models}
\textbf{Choice of the encoder.}
Previous works that aim at learning disentangled representations either focus on randomly initialized RNN-encoders \cite{colombo2021novel,adversarial_removal,coavoux2018privacy} or only use pretrained representations as a feature extractor \cite{ravfogel2020null}. In this work, we choose to fine-tune BERT during training as we believe it to be a more realistic setting.
\\\noindent\textbf{Choice of the baseline models.} We choose to compare our methods against adversarial training from \citet{adversarial_removal,coavoux2018privacy} (model named ADV) and the recently MI bound introduced in \cite{colombo2021novel} (named {MI}) which has been shown to be more controllable than previous MI-based estimators.

\section{Numerical Results}\label{sec:exp_results}

In this section, we gather experimental results for fair classification task. We study our framework when working either with RNN or BERT encoders. The parameter $\lambda$ (see (\ref{eq:problem})) controls the trade-off between success on the main task and disentanglement for all models.

\subsection{Overall Results}\label{ssec:overal_results}
\begin{table}[]
    \centering
     \resizebox{0.45\textwidth}{!}{ \begin{tabular}{cc|ccc|ccc}\hline\hline
           && \multicolumn{3}{c}{\texttt{RNN}} & \multicolumn{2}{c}{\texttt{BERT}} \\\hline
    Dat.      & Loss & $\lambda$& $Y$($\uparrow$) & $S$($\downarrow$) & $\lambda$ &$Y$($\uparrow$) & $S$($\downarrow$)  \\\hline\hline

   \multirow{6}{*}{Sent.}  &  CE  &0.0   &73.2  &68.7  &0.0  &76.2  &76.7  \\
                            &   ADV &1.0   & 71.9 &  56.1       & 0.1 & 74.9 & 72.3 \\
                              & MI &0.1   & 71.6  &  56.3       & 0.1 & 74.5 & 70.3 \\
                            &   W &10     &  \textbf{69.3} & \textbf{50.0}      & 0.01  & \textbf{72.3} & \textbf{54.2} \\
                             &  J   &10   & 70.0  &  54.1        & 10 & 56.7  &56.7 \\
                        &  FR &10 & 57.6 & 52.0         & 10  & 57.4 & 57.4 \\
                     &  MMD &10 & 70.3 & 55.7         & 0.1  & 71.0  & 56.2 \\
                    &  SD &10 & 70.4  &  56.5         & 0.1  &  \textbf{73.8} & \textbf{54.3} \\ \hline\hline
                                                                        
 \multirow{6}{*}{Ment.}     & CE   &0.0  & 77.5  &66.1  &0.0   &81.7  & 79.1  \\
                            &   ADV & 0.1   & 77.0 &   55.4      & 0.1  & 82.2 & 75.3 \\
                              & MI & 10   &  70.0 & 55.7        & 10 & 74.9  & 55.0 \\
                            &   W &10     & \textbf{77.6} & \textbf{50.0}         & 0.01 & 79.0  & 53.0 \\
                             &  J   &10   & 73.4  & 53.3         & 1 & 53.5  & 56.9 \\
                        &  FR &10 & 75.6  &  53.6       &1   & 60.0  & 60.0 \\
                     &  MMD &10 & 77.8 &    58.0    & 0.1 & \textbf{80.0}  & \textbf{52.4} \\
                    &  SD &10 & 77.8  & 56.8            &  0.1 & \textbf{78.4} & \textbf{52.3}\\ \hline\hline
    \end{tabular}}
    \caption{Results on the fair classification task: the main task (higher is better) accuracy corresponds to the column with $Y(\uparrow)$ and $S(\downarrow)$ denotes the sensitive task accuracy (lower is better). CE refers to a classifier trained with CE loss solely ($\lambda=0$ in \eqref{eq:all_loss}).\vspace{-.7cm}}
    \label{tab:table_overall_results}
\end{table}

\begin{figure}[h]
    \centering\vspace{-.2cm}
    \hspace{-0.18cm}
    \subfloat[\centering $Y$ Mention ]{\includegraphics[width=3.75cm]{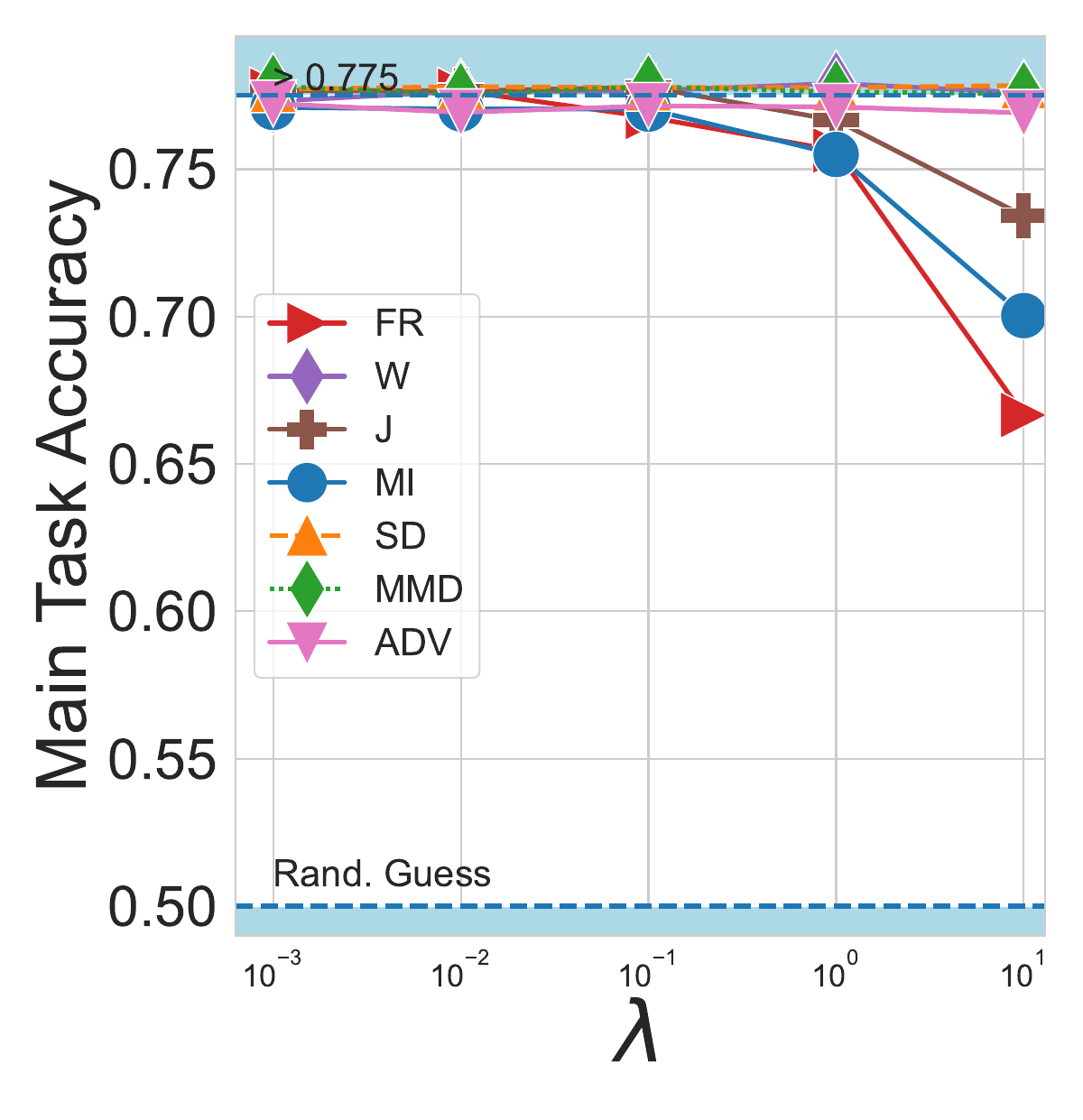}}%
    \subfloat[  \centering  $S$ Mention ]{\includegraphics[width=3.75cm]{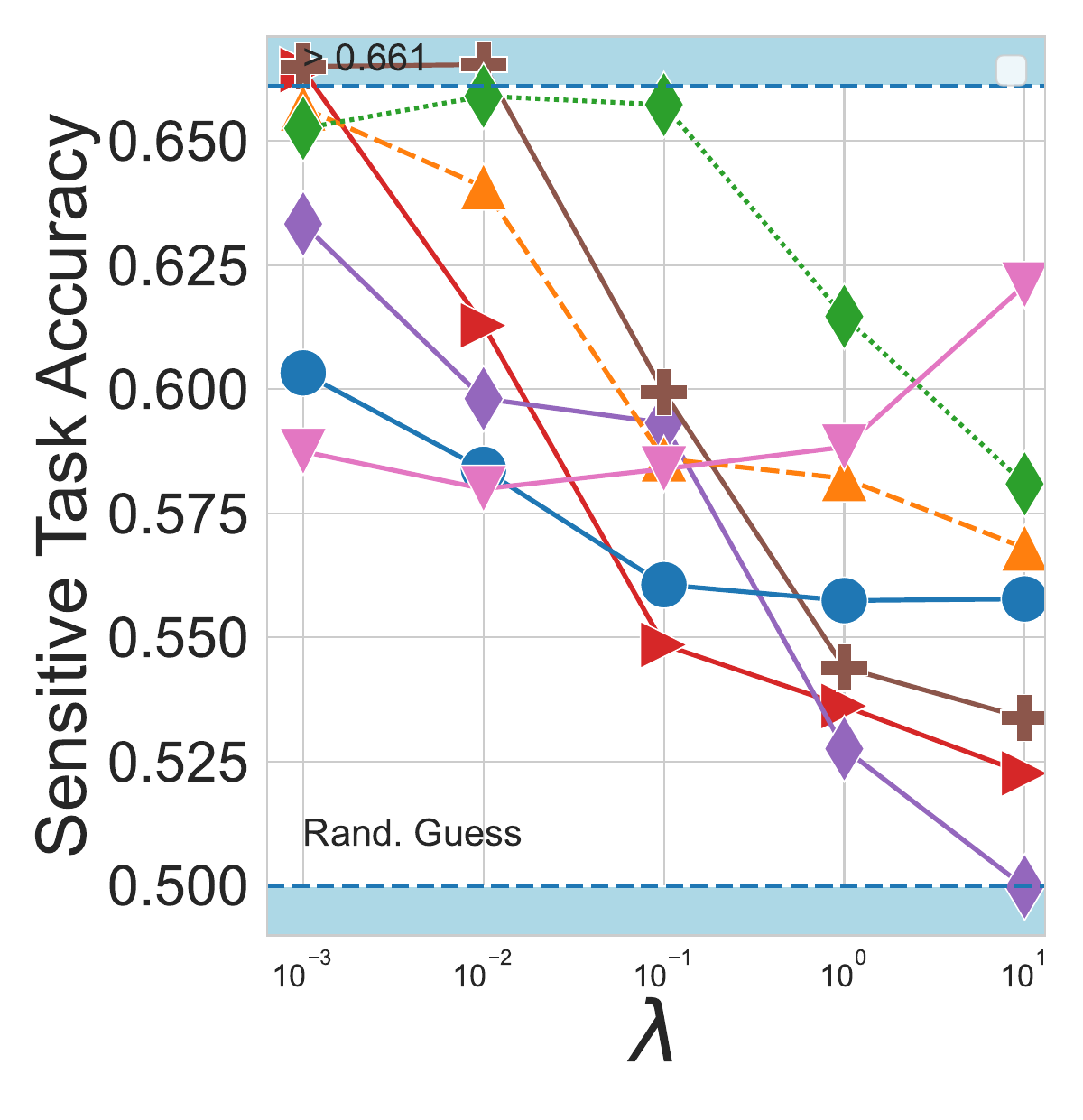}}\\
    \hspace{-0.18cm}
    \subfloat[  \centering  $Y$ Sentiment ]{{\includegraphics[width=3.75cm]{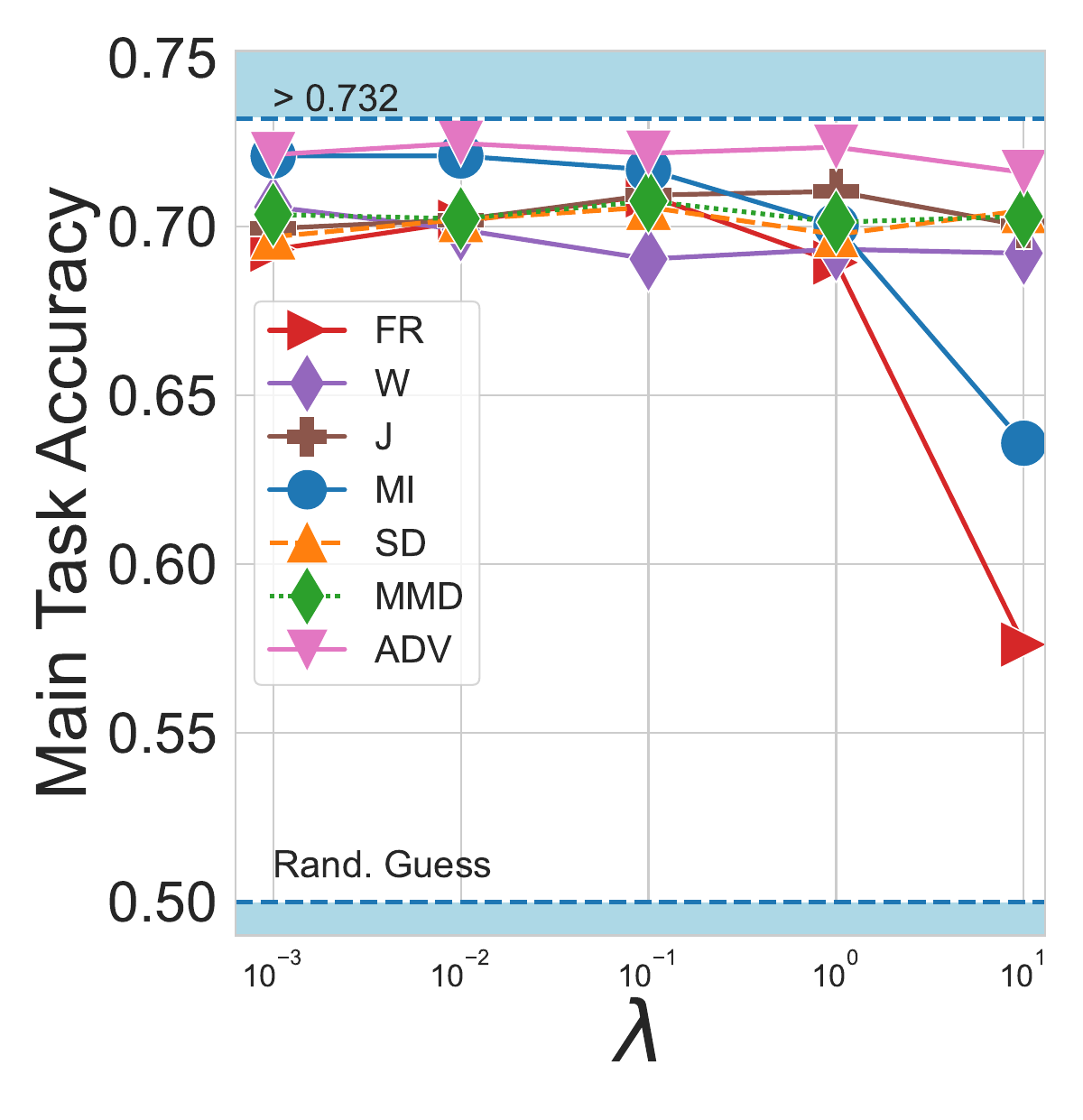} }}%
    \subfloat[  \centering  $S$ Sentiment ]{{\includegraphics[width=3.75cm]{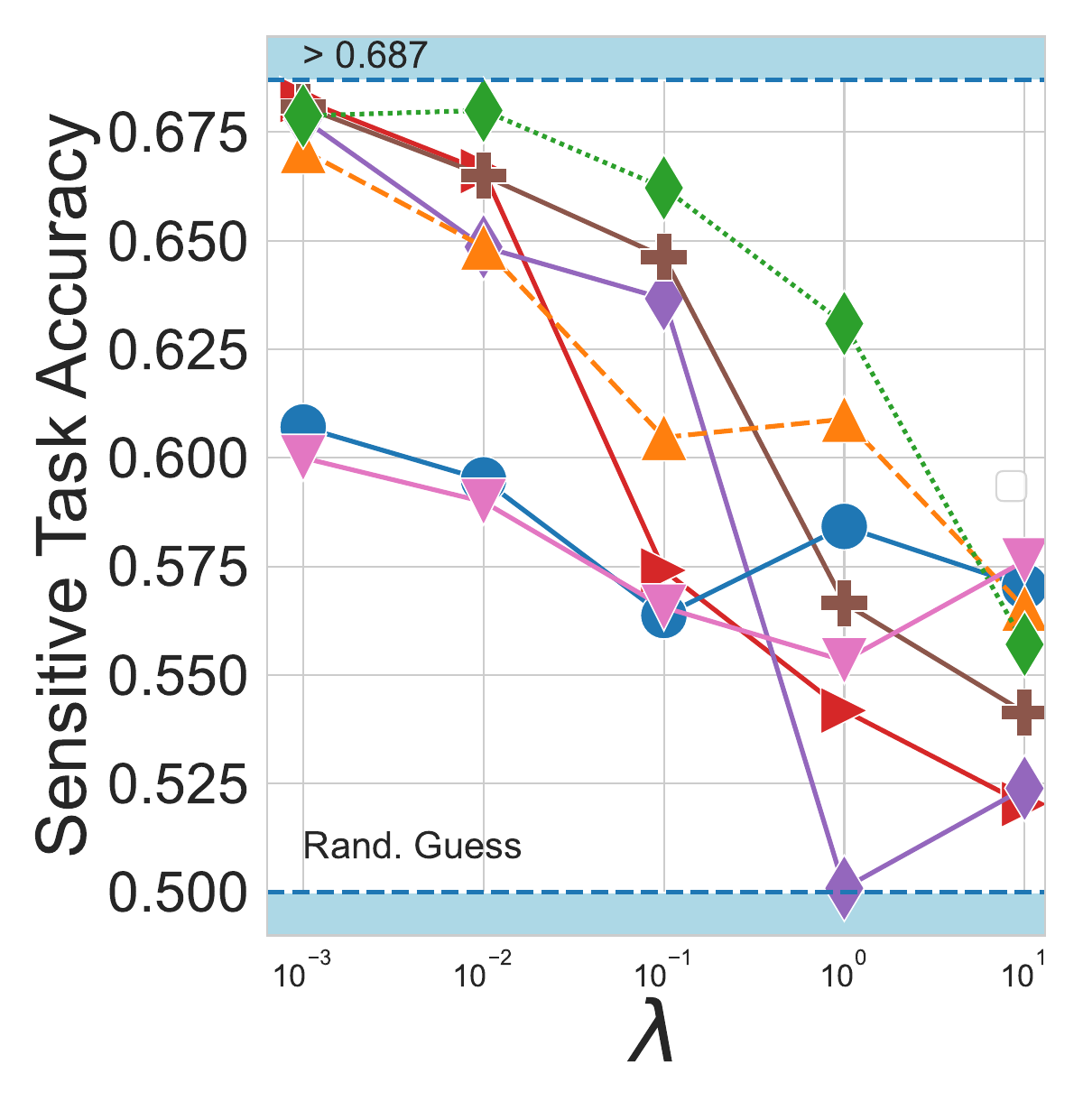} }}%
    \caption{Results on \texttt{DIAL} with RNN. Dash lines correspond to model trained with CE  loss solely (i.e., case $\lambda = 0$ in (\ref{eq:all_loss})). Figures on the left are dedicated to the mention attribute while the one on the rights reports results on the Sentiment attribute. The main task consists in predicting $Y$ thus higher is better. The sensitive task accuracy is obtained by training a classifier to $S$ on the final representation thus an ideal model would reach $50\%$ of accuracy.}%
    \label{fig:fair_classif_rnn_dial}
       \vspace{-.2cm}
\end{figure}

\begin{figure}[h]
    \centering\vspace{-.2cm}
     \hspace{-0.18cm}  
    \subfloat[\centering   $Y$ Mention]{{\includegraphics[width=3.75cm]{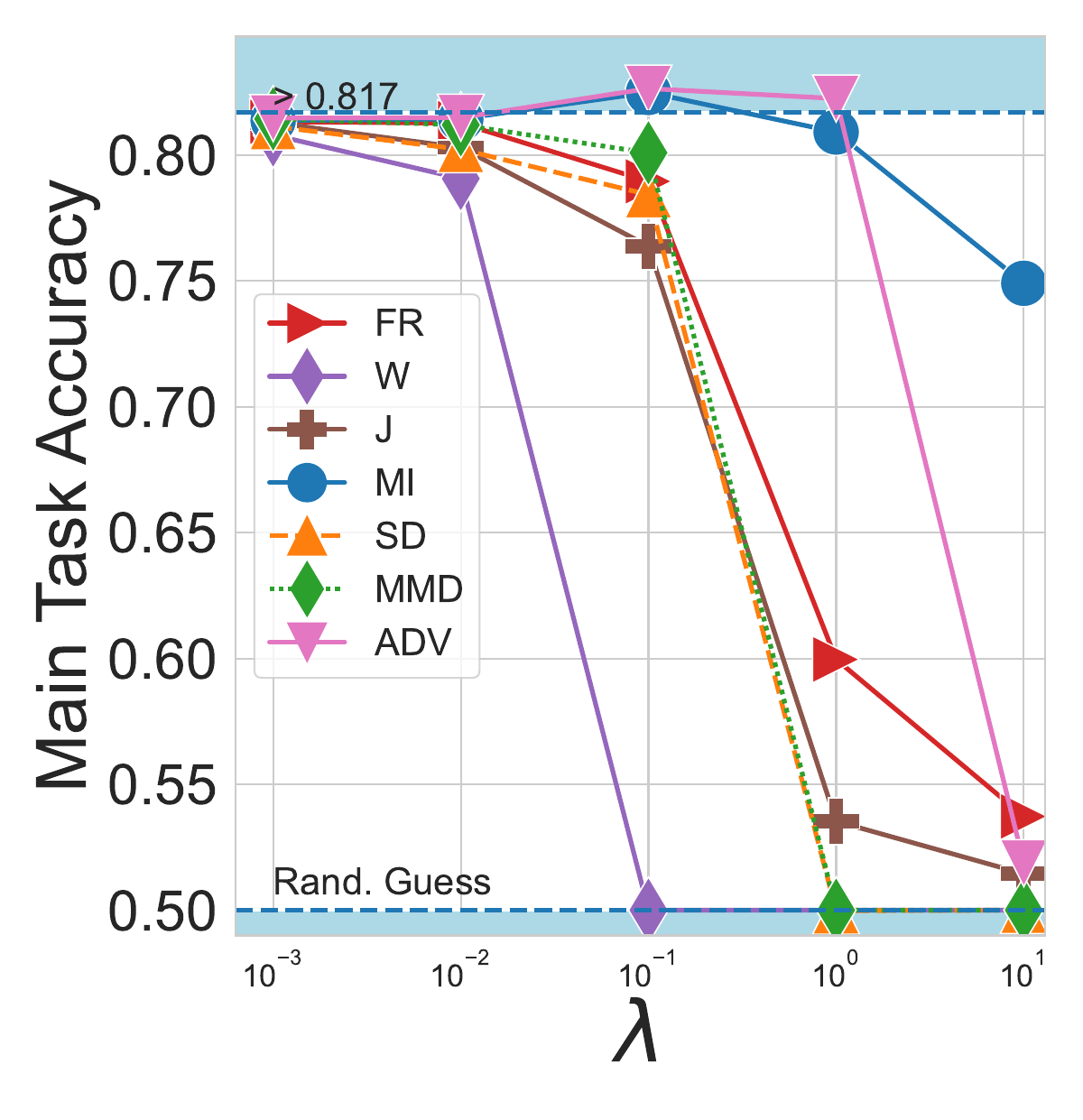} }}%
    \subfloat[\centering   $S$ Mention ]{{\includegraphics[width=3.75cm]{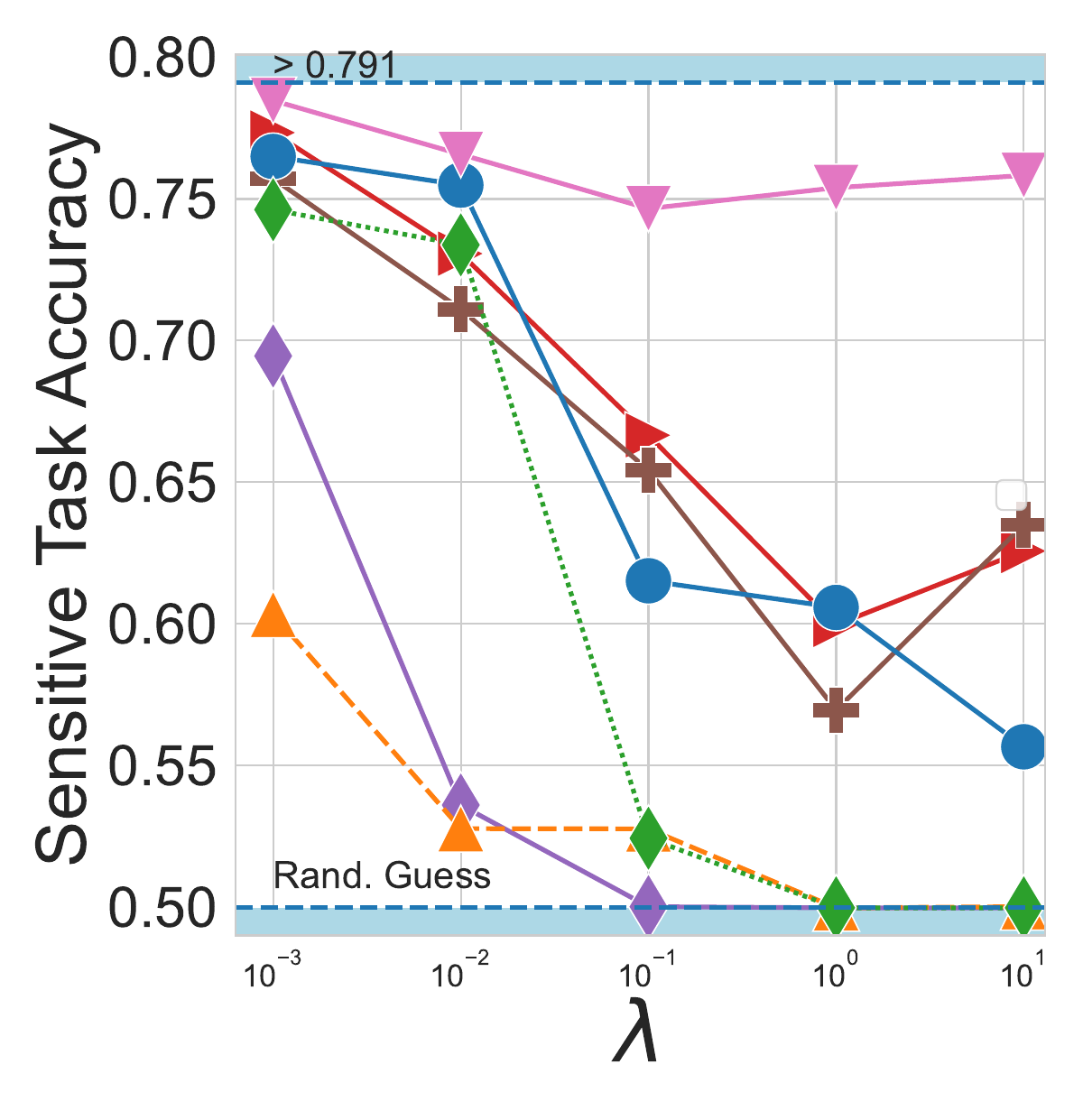} }} \\
    \subfloat[\centering  $Y$ Sentiment ]{{\includegraphics[width=3.75cm]{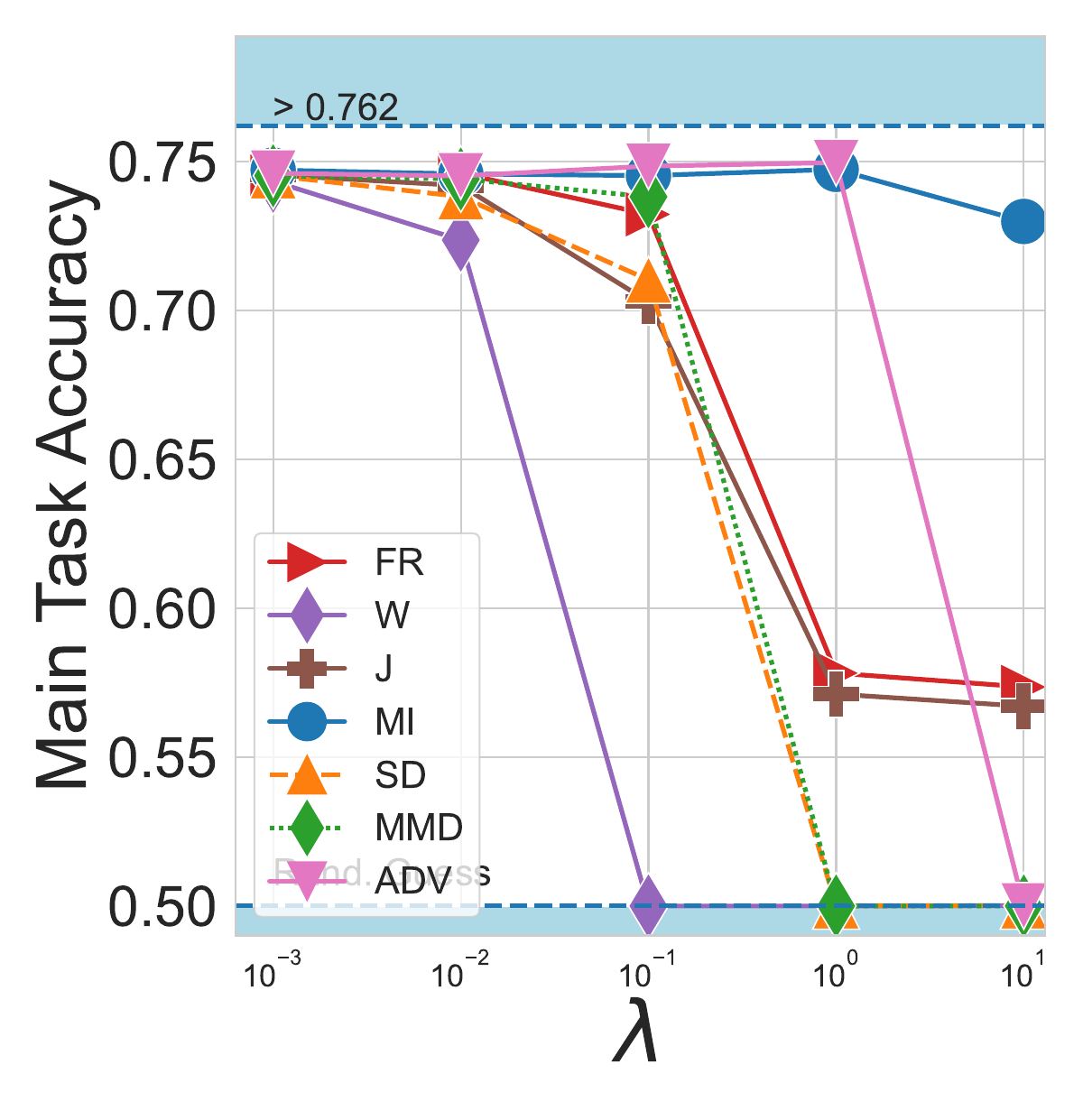} }}%
    \subfloat[\centering  $S$ Sentiment ]{{\includegraphics[width=3.75cm]{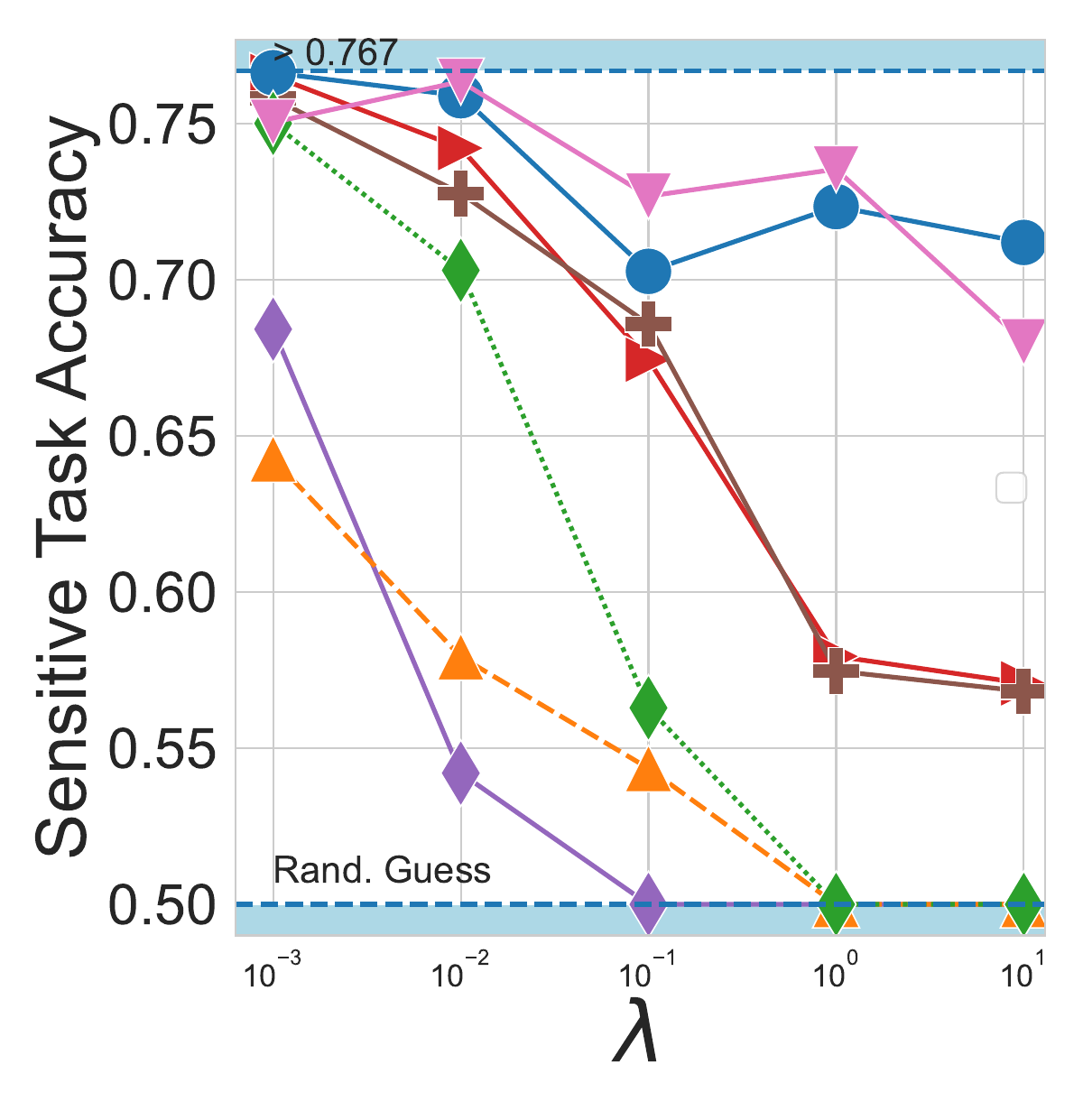} }}%
    \caption{Results on \texttt{DIAL} for mention (left) and sentiment (right) attribute using a pretrained BERT.}%
    \label{fig:fair_classif_bert_dial}%

\end{figure}

\textbf{General observations.} Learning disentangled representations is made more challenging when $S$ and $Y$ are tightly entangled. By comparing Fig.~\ref{fig:fair_classif_rnn_dial} and Fig.~\ref{fig:fair_classif_bert_dial}, we notice that the race label (main task) is easier to disentangled from the sentiment compared to the mention.
\\\noindent\textbf{Randomly initialized RNN encoders.} To allow a fair comparison with previous works, we start by testing our framework with RNN encoders on the \texttt{DIAL} dataset. Results are depicted in Fig.~\ref{fig:fair_classif_rnn_dial}. It is worth mentioning that we are able to observe a similar phenomenon that the one reported in \citet{colombo2021novel}. More specifically, we observe: (i) the adversary degenerates for $\lambda=10$ and does not allow to reach perfectly disentangled representations nor to control the desirable degree of disentanglement; (ii) the {MI}  allows better control over the desirable degree of disentanglement and achieves better-disentangled representations at a reduced cost on the main task accuracy. Fig.~\ref{fig:fair_classif_rnn_dial} shows that the encoder trained using the statistical measures of
similarity--both with and without the multivariate Gaussian assumption--are able to learn disentangled representations. We can also remark that our losses follow an expected behaviour: when $\lambda$ increases, more weight is given to the regularizer, the sensitive task accuracy decreases, thus the representations are more disentangled according to the probing-classifier. Overall, we observe that the {W} regularizer is the best performer with optimal performance for $\lambda=1$ on both attributes. On the other hand, we observe that {FR} and {J} divergence are useful to learn to disentangle the representations but disentangling using these similarity measures comes with a greater cost as compared to {W}. Both {MMD} and {SD} also perform well\footnote{For both losses when $\lambda > 10$ we did not remark any consistent improvements.} and are able to learn disentangled representations with little cost on the main task performance. However, on \texttt{DIAL}, they are not able to learn perfectly disentangled representations. Similar conclusions can be drawn on \texttt{PAN} and results are reported in App. ~\ref{sec:additionnal_results_pan}.

\vspace{1mm}
\noindent\textbf{BERT encoder.} Results of the experiment conducted with BERT encoder are reported in Fig.~\ref{fig:fair_classif_bert_dial}. As expected, we notice that on both tasks the main and the sensitive task accuracy for small values of $\lambda$ is higher than when working with RNN encoders. When training a classifier without disentanglement constraints (i.e., case $\lambda = 0$ in (\ref{eq:all_loss})), which corresponds to the dash lines in Fig.~\ref{fig:fair_classif_rnn_dial} and Fig.~\ref{fig:fair_classif_bert_dial}, we observe that BERT encoder naturally preserves more sensitive information (i.e., measured by the accuracy of the adversary) than randomly initialized encoder. Contrarily to what is usually undertaken in previous works (e.g., \citet{ravfogel2020null}), we allow the gradient to flow in BERT encoder while preforming fine-tuning. We observe a different behavior when compared to previous experiments. Our losses under the Multivariate diagonal Gaussian assumption (i.e., {W}, {J}, {FR} ) can only disentangle the representations at a high cost on the main task (i.e., perfect  disentanglement corresponds to performance on the main task close to a random classifier). When training the encoder with either {SD} or {MMD}, we are able to learn disentangled representations with a limited cost on the main task accuracy: $\lambda=0.1$ achieves good disentanglement with less than 3\% of loss in the main task accuracy. The methods allow little control over the degree of disentanglement and there is a steep transition between light protection with no loss on the main task accuracy and strong protection with discriminative features destruction. 
\vspace{1mm}

\noindent\textbf{Takeaways.} Our new framework relying on statistical Measures of Similarity introduces powerful methods to learn disentangled representations. When working with randomly initialized RNN encoders to learn disentangled representation, we advise relying on {W}. Whereas in presence of pretrained encoders (i.e., BERT), we observe a very different behavior~\footnote{To the best of our knowledge, we are the first to report such a difference in behavior when disentangling attributes with pretrained representations.} and recommend using {SD}. 

\begin{figure*}[h]
    \centering
    \subfloat[\centering   $Y$ RNN]{{\includegraphics[width=3.75cm]{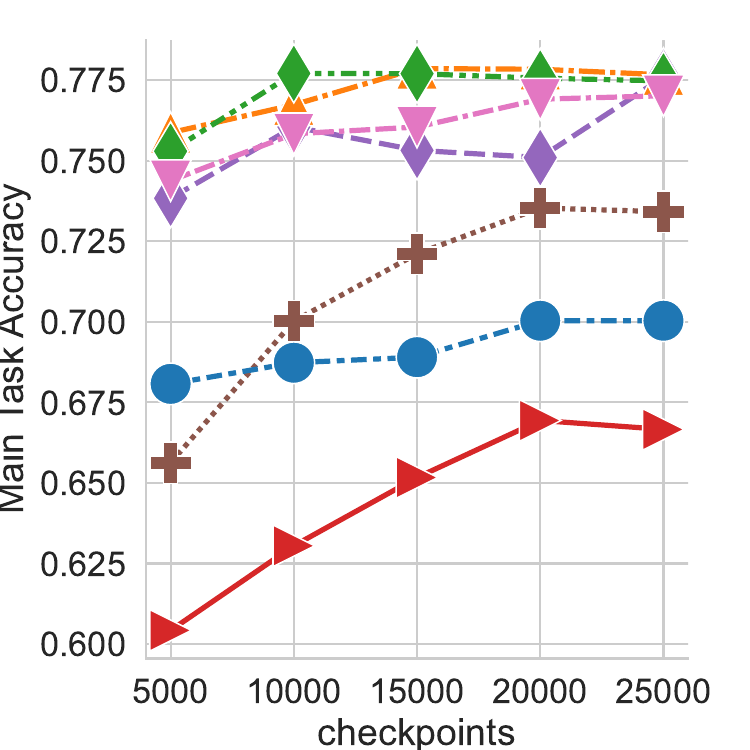} }}%
    \subfloat[\centering $S$ RNN ]{{\includegraphics[width=3.75cm]{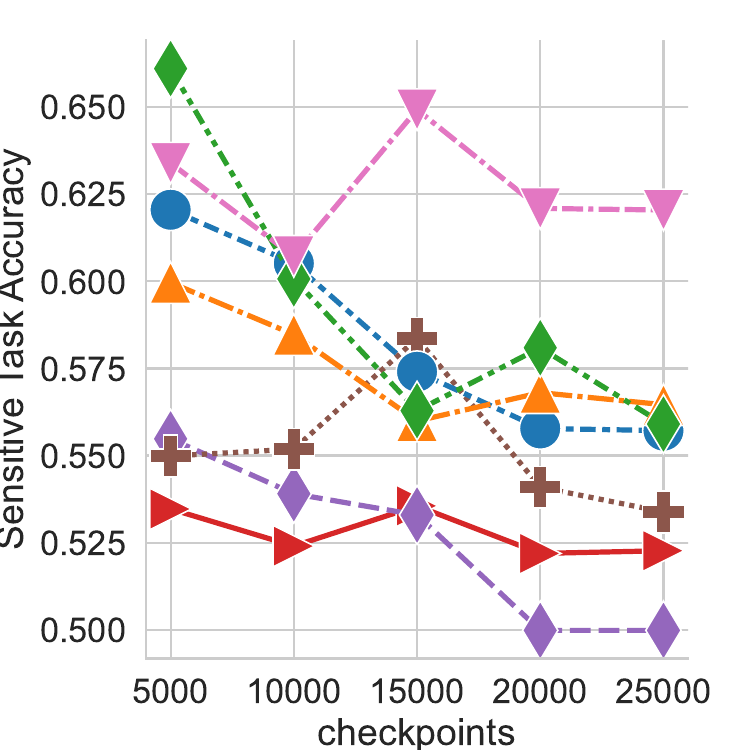} }}
    \subfloat[\centering   $Y$ BERT  ]{{\includegraphics[width=3.75cm]{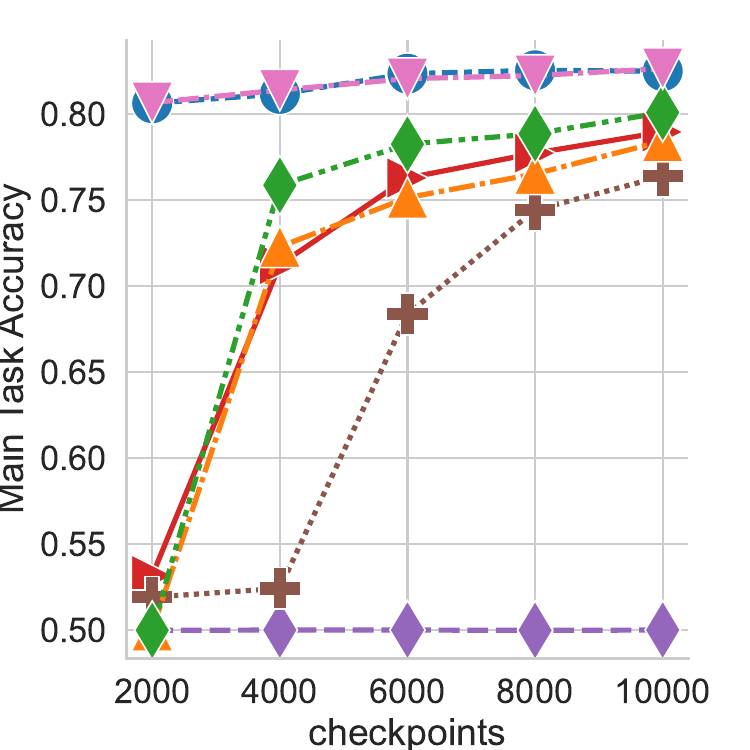} }}%
    \subfloat[\centering   $S$ BERT ]{{\includegraphics[width=4.75cm]{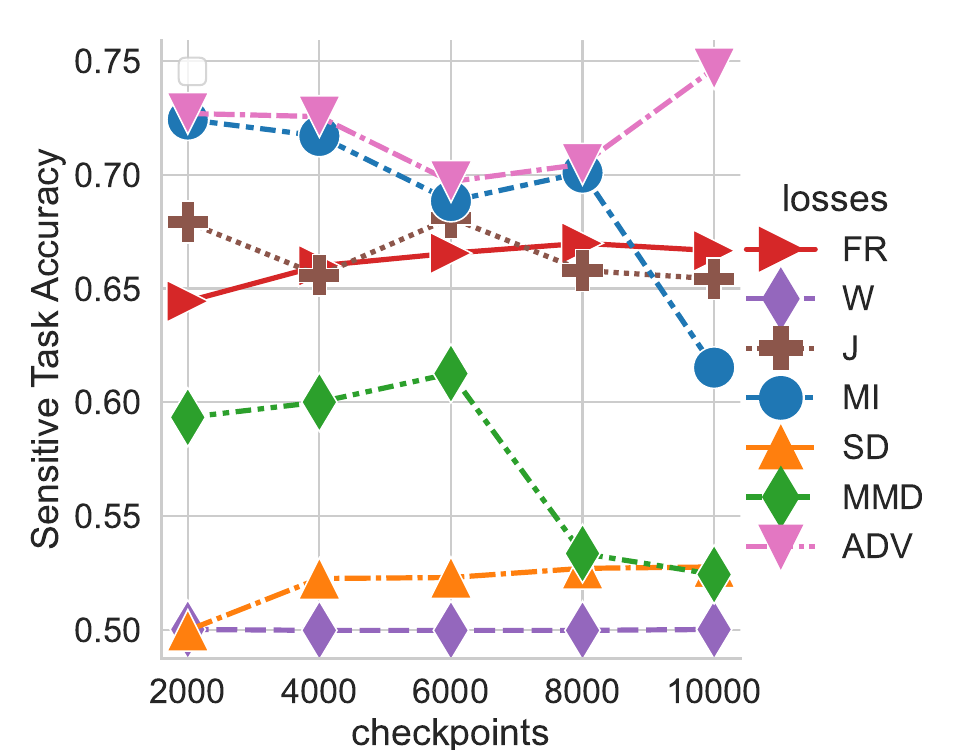} }}%
   \caption{Training Dynamic on \texttt{DIAL} for the mention label with using RNN (left $\lambda = 10$ ) and BERT (right $\lambda = 0.1$) encoders.}%
    \label{fig:training_dynamic}
       \vspace{-.5cm}
\end{figure*}

\begin{table}[h]
    \centering
  \resizebox{0.45\textwidth}{!}{ \begin{tabular}{lcccc}\hline\hline
 & Method & \# params. &  1 upd. & 1 epoch. \\\hline\hline
  \multirow{7}{*}{\rotatebox[origin=c]{90}{RNN}
 }  &     ADV    &  \result{2220}{-0.6\%} &   0.11 & \result{551}{-17\%}    \\
    &   MI   & 2234 &    0.13 & 663      \\\cline{2-5}
    & {FR}  &    \multirow{5}{*}{\result{2206}{-1.3\%}} & 0.10 & \result{508}{-23\%}         \\
    & W &   &  0.10 & \result{509}{-23\%}    \\
    & J &   &  0.10 & \result{507}{-23\%}       \\
    & {MMD} &   &   0.10 & \result{520}{-22\%}      \\
    & {SD} &    & 0.10 &  \result{544}{-18\%}  \\\hline\hline
     & Method & \# params. &  1 upd. & 1 epoch.  \\\hline\hline
\multirow{7}{*}{\rotatebox[origin=c]{90}{BERT}}&                  ADV    & \result{109576}{-0.01\%} &  0.48 & \result{2424}{-10\%}     \\
   &  MI   &  109591 &  0.55 & 2689  \\\cline{2-5}
  &   FR &  \multirow{5}{*}{\result{109576}{-0.03\%} }   &  0.47 & \result{2290}{-14\%}      \\
  &    W &   &  0.47 & \result{2290}{-14\%}    \\
  &  J &   &  0.47 & \result{2307}{-14\%}    \\
    &    {MMD}  &   &  0.48 &   \result{2323}{-14\%}     \\
    &            {SD} &    & 0.48 & \result{2347}{-13\%}   \\\hline\hline
     \end{tabular}}
    \caption{Speed and number of model parameters (given in thousand)  when working with \texttt{DIAL}. The runtime for 1 gradient update (denoted 1 upd.) or for 1 epoch is given for a batch of 64 when running our models on a single NVIDIA-V100. The relative improvements (in \%) are given with respect to the {MI}  model, which is our strongest baseline.}\vspace{-0.5cm}
    \label{tab:training_time_and_parameters}
\end{table}

\begin{figure}[h]\vspace{-.5cm}
    \subfloat[ RNN]{{\includegraphics[width=3.75cm]{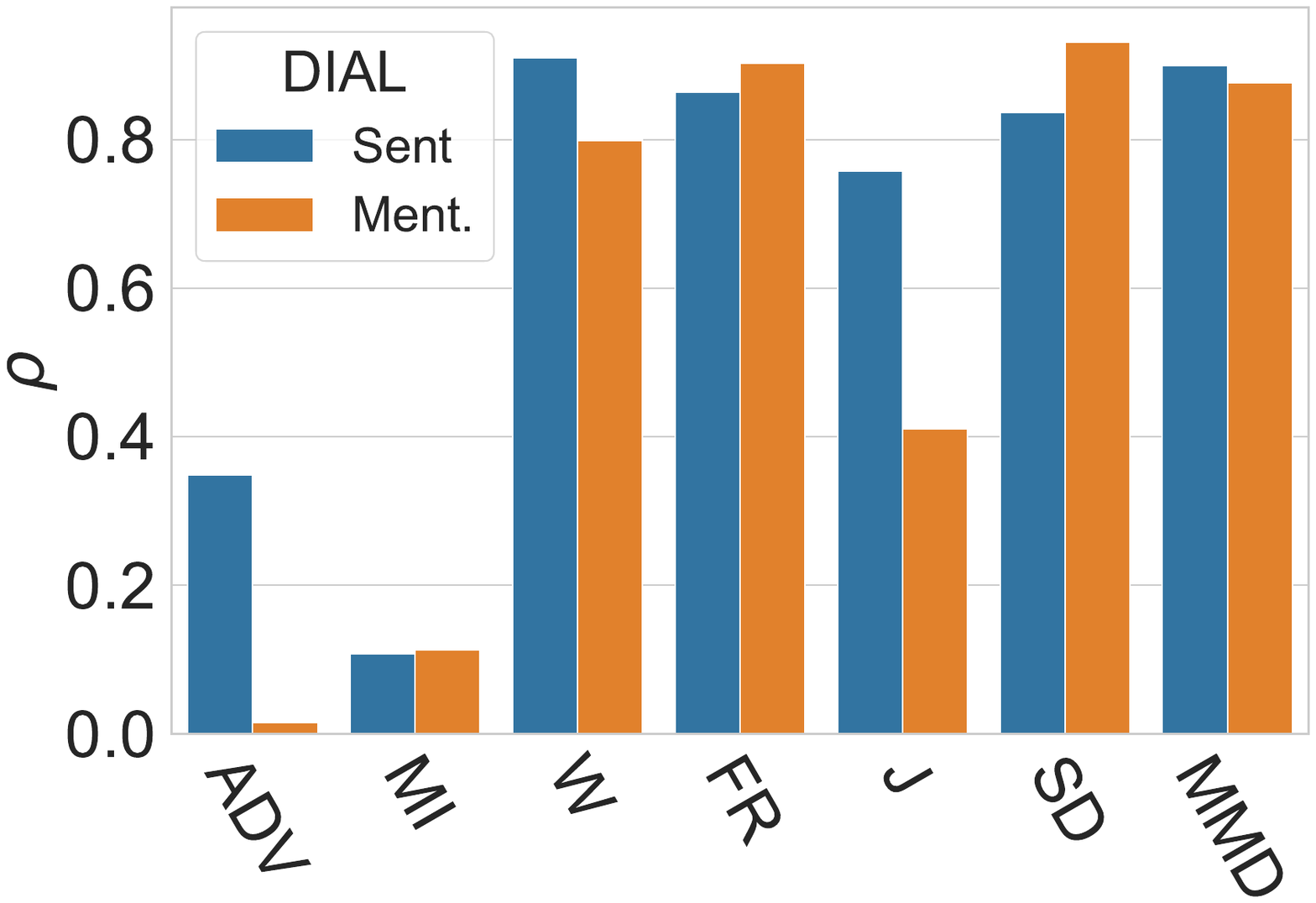}}}%
    \subfloat[ BERT  ]{{\includegraphics[width=3.75cm]{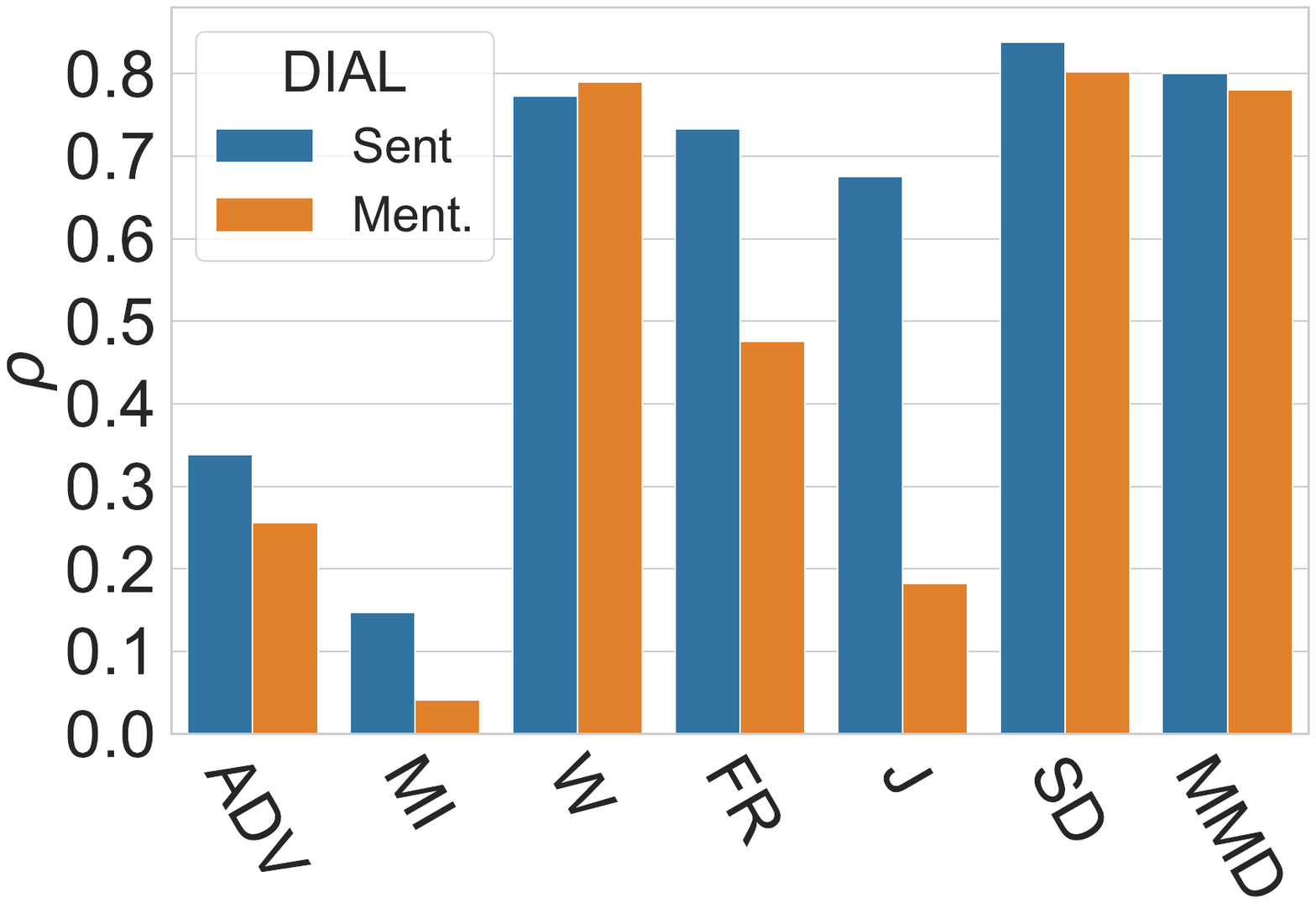}}}\vspace{-.3cm}
    \caption{ Absolute Pearson correlation between the values of $R$ and the sensitive task accuracy.}%
    \label{fig:correlation}\vspace{-.5cm}
\end{figure}

\subsection{Speed Gain and Parameter Reduction}
We report in Table \ref{tab:training_time_and_parameters} the training time and the number of parameters of each method. The reduced number of parameters brought by our method is marginal, however getting rid of these parameters is crucial. Indeed, they require a nested loop and require a fined selection of the hyperparameters which complexify the global system dynamic. 
\\\noindent\textbf{Takeaways.} Contrarily to MI or Adversarial based regularizer that are difficult (or even prohibitive) to be implemented on large-scale datasets, our framework is simpler and consistently faster which makes it a better candidate when working with large-scale datasets.

\section{Further Analysis}
Results presented in Section \ref{ssec:overal_results} have shown a different behaviour for RNN and BERT based encoders and, for different measures of similarity. Here, we aim at understanding of this phenomena.

\subsection{Training Dynamic}
In the previous section, we examine the change of the measures during the training.
\\\noindent\textbf{Takeaways.} When using a RNN encoder, the system is able to maximize the main task accuracy while jointly minimizing most of the similarity measures. For BERT where the model is more complex, for measures relying on the diagonal gaussian multivariate assumption either the disentanglement \emph{plateau} (e.g., {FR} or {J})  or the system fails to learn discriminative features and perform poorly on the main task (e.g.,  W). When combined with BERT both SD and MMD can achieve high main task accuracy while protecting the sensitive attribute. 

\subsection{Correlation Analysis}
In this experiment, we investigate how predictive of the disentanglement is each similarity measure, i.e., does a lower value of similarity measure indicates better disentangled representations? We gather for both the mention and sentiment attribute 5 checkpoints per model (i.e., each regularizer and each value of $\lambda$ corresponds to one model). For each RNN model, we select one checkpoint after $5\mathrm{k},10\mathrm{k},15\mathrm{k},20\mathrm{k},25\mathrm{k}$ gradient updates, and for BERT we select one checkpoint after $2\mathrm{k},4\mathrm{k},6\mathrm{k},8\mathrm{k},10\mathrm{k}$ gradient updates to obtain the same number of models. For each  type of loss, we ended with 50 models. For each model and each checkpoint, we train an adversary, compute the sensitive task accuracy and evaluate the Pearson correlation between the sensitive task accuracy and the corresponding similarity measure. Results are presented  in Fig.~\ref{fig:correlation}.
\vspace{1mm}

\noindent\textbf{Takeaways.} Both ADV and {MI} poorly are correlated with the degree of disentanglement of the learned representations. We find this result not surprising at light of the findings of \citet{adv_classif_fair_1} and \citet{song2019understanding}. All our losses achieve high correlation ($\rho \geq 78$) except for {J} in the mention task with both encoders, and the {FR}  with BERT on the mention task that achieves medium/low correlation. We believe, that the high correlation showcases the validity of the proposed approaches.

\section{Summary and Concluding Remarks}
We have introduced a new framework for learning disentangled representations which is faster to train, easier to tune and achieves better results than adversarial or MI-based methods. Our experiments on the fair classification task show that for RNN encoders, our methods relying on the closed-form of similarity measures under a multivariate Gaussian assumption can achieve perfectly disentangled representations with little cost on the main tasks (e.g. using Wasserstein). On BERT representations, our experiments show that the Sinkhorn divergence should be preferred. It can achieves almost perfect disentanglement at little cost but allows for fewer control over the degree of disentanglement.  

\section*{Acknowledgments}
This work was also granted access to the HPC resources of IDRIS under the allocation 2021-AP010611665 as well as under the project 2021-101838 made by GENCI. This work has been supported by the project PSPC AIDA: 2019-PSPC-09 funded by BPI-France. 

\bibliography{anthology,custom}
\bibliographystyle{acl_natbib}
\newpage
\clearpage
\appendix

\section{Additional details on Statistical Measures of Similarity}\label{sec:all_metric_detailled}

It is the purpose of this part to recall additional details on similarity measures defined in the core paper.
\subsection{Univariate Fisher-Rao distance}\label{subsec:FR}

Here, we recall the definition of the univariate Fisher-Rao distance used in the Section \ref{ssec:similarity_function}. Let $\mathbb{Q}_1,\mathbb{Q}_2$ be two univariate probability distributions with mean $m_1,m_2 \in \mathbb{R}$ and standard deviation $\sigma_1,\sigma_2 \in \mathbb{R}$. Thus, the univariate Fisher-Rao distance between the tuples $(m_1,\sigma_1)$ and $(m_2,\sigma_2)$ denoted by  $d_{\text{FR}}$ is defined as:

{\scriptsize
\begin{equation*}\label{fisheruniv}
\sqrt{2}\log \frac{\sqrt{\frac{(m_{1}-m_{2})^2}{2} + (\sigma_{1}+\sigma_{2})^2} + \sqrt{\frac{(m_{1}-m_{2})^2}{2} + (\sigma_{1}-\sigma_{2})^2}}{\sqrt{\frac{(m_{1}-m_{2})^2}{2} + (\sigma_{1}+\sigma_{2})^2} - \sqrt{\frac{(m_{1}-m_{2})^2}{2} + (\sigma_{1}-\sigma_{2})^2}}. 
\end{equation*}
}

\subsection{Empirical versions of Statistical Measures of Similarity}\label{subsec:emp}

Our experimental setting involves the tuple $\left( X^i, S^i, Y^i \right)_{i=1}^n$
where $X^1,\ldots,X^n$ is a sample of texts drawn from the random variable $X\in \mathcal{X}$,  $S^1,\ldots,S^n$ is a sample of binary random variables corresponding to the sensitive attribute and $Y^1,\ldots,Y^n$ is a sample of binary random variables coming from the classification task. Considering the embedding function $f_{\theta}$, we denote by $Z^1,\ldots,Z^n$ the embedding sample such that $Z^i=f_{\theta}(X^i)$ for every $1\leq i\leq n$. Assume that $\{ i_1,\ldots,i_{n_0}\}$ and $\{j_1,\ldots,i_{n_1}\}$ are two subsets of $\{1,\ldots, n \}$ such that $S_{i_k}=0$ and $S_{j_l}=1$ for every $1\leq k\leq n_0$ and $1\leq l\leq n_1$. The empirical versions of the conditional measures $\mathbb{P}_0,\mathbb{P}_1$ are given by 

\begin{equation*}
    \hat{\mathbb{P}}_0=\frac{1}{n_0} \sum_{i \in \{i_1,\ldots,i_{n_0} \} } \hspace{-0.1cm}\delta_{Z^i}, \;   \hat{\mathbb{P}}_1=\frac{1}{n_1} \sum_{j \in \{j_1,\ldots,j_{n_1} \} } \hspace{-0.1cm}\delta_{Z^j}.
\end{equation*}
In practice, distances recalled in Section ~\ref{ssec:similarity_function} are computed between $\hat{\mathbb{P}}_0$ and $\hat{\mathbb{P}}_1$ leading to the following distances. 

\vspace{0.2cm}

\noindent \textbf{Maximum Mean Discrepancy.} The MMD is defined as:

\begin{align*}
    \text{MMD}(\hat{\mathbb{P}}_0,\hat{\mathbb{P}}_1) &= \frac{1}{n_0(n_0-1)} \sum_{\substack{i,k \in \{ i_1,\ldots,i_{n_0}\} \\ i \neq k  }} \hspace{-0.4cm}k(Z^i,Z^k) 
    \\& + \frac{1}{n_1(n_1-1)} \sum_{\substack{j,l \in \{ j_1,\ldots,j_{n_1}\} \\ j \neq l  }} \hspace{-0.4cm}k(Z^j,Z^l) 
        \\& - \frac{2}{n_0n_1} \sum_{\substack{i \in \{ i_1,\ldots,i_{n_0}\} \\ j \in \{ j_1,\ldots,j_{n_1}\}  }} \hspace{-0.4cm}k(Z^i,Z^j) .
\end{align*}

\vspace{0.3cm}

\noindent \textbf{Sinkhorn divergence.} Let  $\mathbf{1}_{n_0},\mathbf{1}_{n_1}$ denote the vectors of one with size $n_0,n_1$ respectively. Let $\mathcal{U}(\hat{\mathbb{P}}_0, \hat{\mathbb{P}}_1)= \{\Pi \in \mathbb{R}^{n_0\times n_1}  \; |\; \; \Pi \mathbf{1}_{n_1} = \mathbf{1}_{n_0}/n_0, \; \; \Pi^\top \mathbf{1}_{n_0} = \mathbf{1}_{n_1}/n_1  \}$  be the set of joint probability distributions with marginals $\hat{\mathbb{P}}_0$ and $\hat{\mathbb{P}}_1$ where $/$ is the element-wise division. The Sinkhorn approximation of the 1-Wasserstein distance, denoted by $\text{W}_{\varepsilon}( \hat{\mathbb{P}}_0, \hat{\mathbb{P}}_1)$,  is defined as the following optimization problem:

{\small 
\begin{equation*}
 \underset{ \Pi~\in~\mathcal{U}(\hat{\mathbb{P}}_0, \hat{\mathbb{P}}_1)}{\min}  \sum_{\substack{i \in \{i_1,\ldots, i_{n_0} \} \\ j \in \{j_1,\ldots, j_{n_1} \}}} \hspace{-0.2cm}\Pi_{i,j} \text{D}_{i,j}  + \varepsilon \hspace{-0.3cm} \sum_{\substack{i \in \{i_1,\ldots, i_{n_0} \} \\ j \in \{j_1,\ldots, j_{n_1} \}}} \hspace{-0.2cm} \Pi_{i,j} \log(\Pi_{i,j}),
\end{equation*}
}
\noindent where $\text{D}_{i,j}$ is the euclidean distance between $Z^i$ and $Z^j$. We limit ourselves to the 1-Wasserstein for the sake of place. The Sinkhorn divergence is then:

\begin{equation*}
    \text{{SD}}_{\varepsilon} (\hat{\mathbb{P}}_0,\hat{\mathbb{P}}_1)= \text{W}_{\varepsilon}(\hat{\mathbb{P}}_0,\hat{\mathbb{P}}_1) - \frac{1}{2}\sum_{i=0}^{1}\text{W}_{\varepsilon}(\hat{\mathbb{P}}_i,\hat{\mathbb{P}}_i).
\end{equation*}

\noindent It is worth noticing that a robust version of the Wasserstein distance can be found in \cite{staerman2020ot} (see also \cite{dr_distance}).

\noindent \textbf{Fisher-Rao distance.} The Fisher-Rao distance is defined as

\begin{align*}
    \text{FR}(\hat{\mathbb{P}}_0^{\hat{p}_0}, \hat{\mathbb{P}}_1^{\hat{p}_1})= \sqrt{\sum_{j=1}^{d} \left[d_{\text{FR}}(\hat{p}_{0,j},\hat{p}_{1,j})\right]^2},
\end{align*}
\noindent where $d_{\text{FR}}(\hat{p}_{0,j},\hat{p}_{1,j})$ is defined as in Section ~\ref{fisheruniv}, and $m_1,m_2$ and $\sigma_1,\sigma_2$ are replaced by $\hat{\mu}_{0,j},\hat{\mu}_{1,j}$ and $\hat{\sigma}_{0,j},\hat{\sigma}_{1,j}$ the classical (univariate) unbiased mean and standard deviation estimators respectively.

\noindent \textbf{Jeffrey divergence.} Let $(\hat{\mu}_0,\hat{\Sigma}_0)$ and $(\hat{\mu}_1,\hat{\Sigma}_1)$ be the mean and the covariance matrix estimators of the samples $Z^{i_1},\ldots,Z^{i_{n_0}}$ and $Z^{j_1},\ldots,Z^{j_{n_1}}$ respectively. Jeffrey divergence--under the multivariate Gaussian assumption--boils down to:

\begin{align*}\label{eq:kl}
\log \frac{|\hat{\Sigma}_0|}{|\hat{\Sigma}_1|}\hspace{-.05cm}-\hspace{-.1cm}d\hspace{-.1cm}+\hspace{-.05cm}\text{Tr}(\hat{\Sigma}_0^{-1}\hat{\Sigma}_1)  \hspace{-.1cm}+ \hspace{-.1cm}(\hat{\mu}_0\hspace{-.1cm}-\hspace{-.1cm}\hat{\mu}_1)^{T} \hspace{-.05cm}\hat{\Sigma}_0^{-1}\hspace{-.05cm}(\hat{\mu}_0\hspace{-.1cm}-\hspace{-.1cm}\hat{\mu}_1).
\end{align*}

\noindent Furthermore, under the multivariate Gaussian assumption, the Wasserstein distance writes as follows:

\begin{equation*}
    \text{W}(\hat{\mathbb{P}}_0, \hat{\mathbb{P}}_1)\hspace{-.1cm}=\hspace{-.1cm}\|\hat{\mu}_0 -\hat{\mu}_1\|^{2}\hspace{-.05cm}+ \hspace{-.05cm}\mathrm{Tr}\hspace{-.05cm}\left(\hat{\Sigma}_0 \hspace{-.1cm}+\hspace{-.1cm}\hat{\Sigma}_1\hspace{-.1cm}-\hspace{-.1cm}2(\hat{\Sigma}_0 \hat{\Sigma}_1 )^{1/2}\right)
\end{equation*}

\section{Algorithm}\label{sec:algo}

The algorithm we propose to compute (\ref{eq:problem}) involves a simple training loop and is described in Algorithm ~\ref{alg:m_PARED}.

\begin{algorithm}
  \caption{Training Algorithm
    \label{alg:m_PARED}}
  \begin{algorithmic}[1]
   \\\textsc{Input} $\mathcal{D}=\{(x_{j},s_{j},y_{j}), \forall j \in [1,n]\}$, $\theta$
weights of the encoder network, $\phi$ parameters of the main classifier.

\\ \textsc{Initialize} parameters $\theta$, $\phi$.
\While{$(\theta,\phi)$ not converged}  \Comment{Single loop}
\State Sample a batch $\mathcal{B}$ from $\mathcal{D}$
\State Compute CE \Comment{Classification Loss}
\State Compute $\text{SM}$ \Comment{Disentanglement Loss}
\State Update $\theta, \phi$ with $\mathcal{B}$ using AdamW.
\EndWhile
\\ \textsc{Output} Encoder and main classifier weights ${\theta,\phi}$
  \end{algorithmic}
\end{algorithm}
\section{Additional Results}\label{sec:additionnal_results}
In this section, we gather additional experimental results.

\subsection{Results on \texttt{PAN}}\label{sec:additionnal_results_pan}

\begin{table}[]
    \centering
     \resizebox{0.45\textwidth}{!}{ \begin{tabular}{cc|ccc|ccc}\hline\hline
           && \multicolumn{3}{c}{\texttt{RNN}} & \multicolumn{2}{c}{\texttt{BERT}} \\\hline
    Dat.      & Loss & $\lambda$& $Y$($\uparrow$) & $S$($\downarrow$) & $\lambda$ &$Y$($\uparrow$) & $S$($\downarrow$)  \\\hline\hline

   \multirow{6}{*}{Age}  &  CE  &0.0   &85.7  &60.0  &0.0  &87.0  &65.0  \\
                            &   ADV & 1 & 82.5 & 57.1 & 0.01 &87.0 & 60.0 \\
                              & MI  & 0.1 & 81.9 & 56.7 & 0.1 & 86.2 & 62.0 \\
                            &   W  & 10  & \textbf{82.9} & \textbf{50.0} &  0.01 & 84.3 & 52.1  \\
                             &  J   & 1 & 81.3 & 53.8 & 1 & 66.3 & 57.5 \\
                        &  FR  & 10 & 63.3 & 50.0 &10 & 64.4 & 56.5 \\
                     &  MMD  & 10 & 83.3 & 50.1 & 0.1 & 85.2 & 54.4 \\
                    &  SD  & 10 & 80.0 & 50.0  & 0.1 & \textbf{80.2} & \textbf{52.4} \\ \hline\hline
                                                                        
 \multirow{6}{*}{Gender}     & CE   &0.0  & 85.7  &59.1  &0.0   &87.0  & 65.0  \\
                            &   ADV & 0.1 & 77.0 & 55.4& 0.01 & 87.3 & 61.3 \\
                              & MI  & 10 & 70.0  & 55.7 & 0.1& 86.8 & 63.9 \\
                            &   W  & 10 & \textbf{77.6} & \textbf{50.0} & 0.01 & 83.7 & 51.7 \\
                             &  J & 10  & 73.4 & 53.3  & 1 & 62.7 & 54.3 \\
                        &  FR  & 1 & 75.6 & 53.6 & 1  & 62.1 & 58.1 \\
                     &  MMD & 10 & 77.8 & 56.8 & 0.1 &\textbf{85.7} & \textbf{51.4}\\
                    &  SD &10 &77.5 & 58.0 & 0.1 &80.7 &51.7 \\ \hline\hline
    \end{tabular}}
    \caption{Results on the fair classification task: the main task (higher is better) accuracy is correspond to the column with $Y(\uparrow)$ and $S\downarrow$ denotes the sensitive task accuracy respectively (lower is better). $\lambda$ (see Eq.~\ref{eq:problem} control the trade-off between success on the main task and disentanglement.}
    \label{tab:table_overall_results}
\end{table}

\begin{figure}%
    \centering
    \subfloat[\centering $Y$ RNN \texttt{PAN}-Age]{\includegraphics[width=4cm]{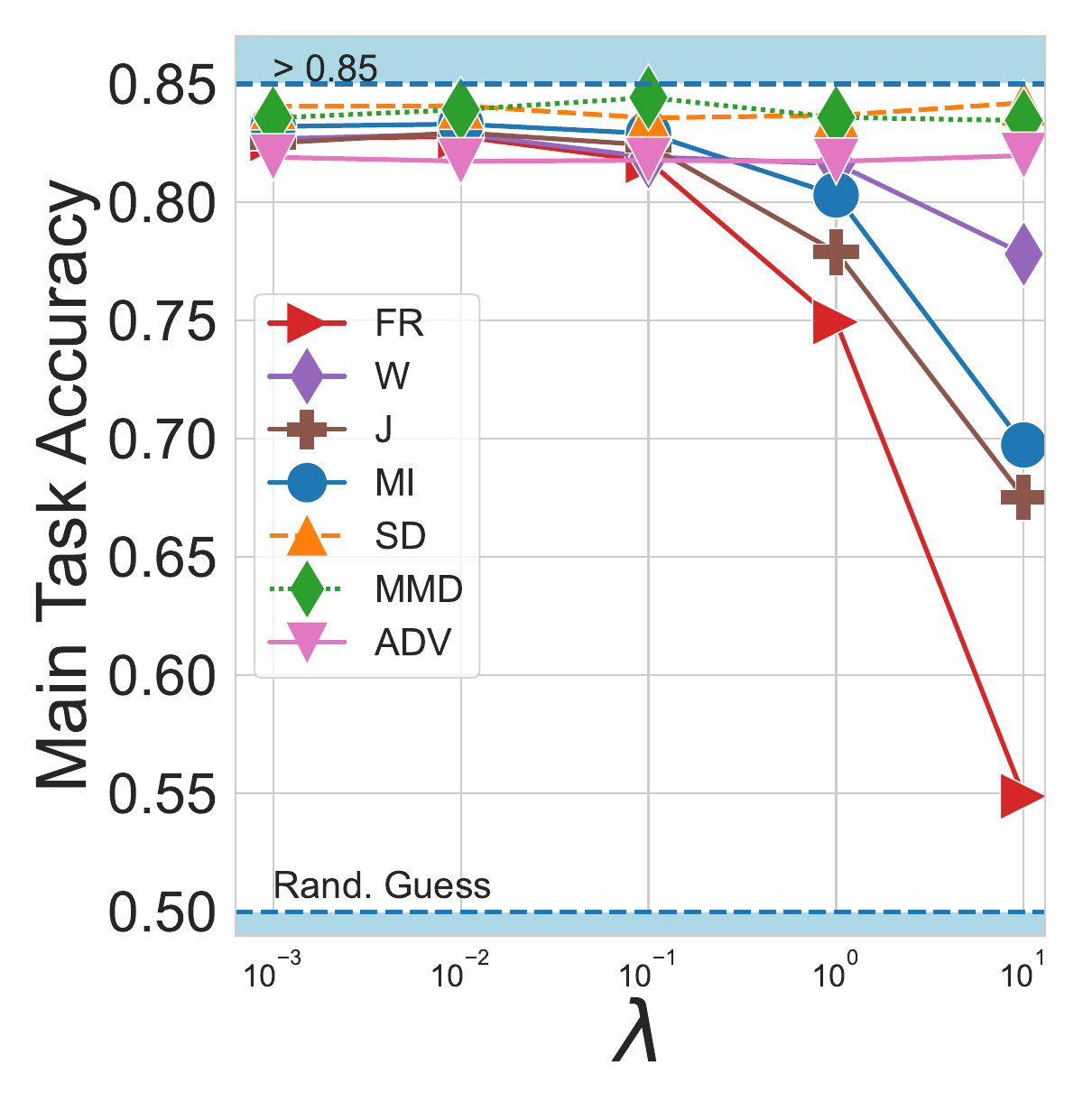}}%
    \subfloat[\centering $S$ RNN \texttt{PAN}-Age ]{\includegraphics[width=4cm]{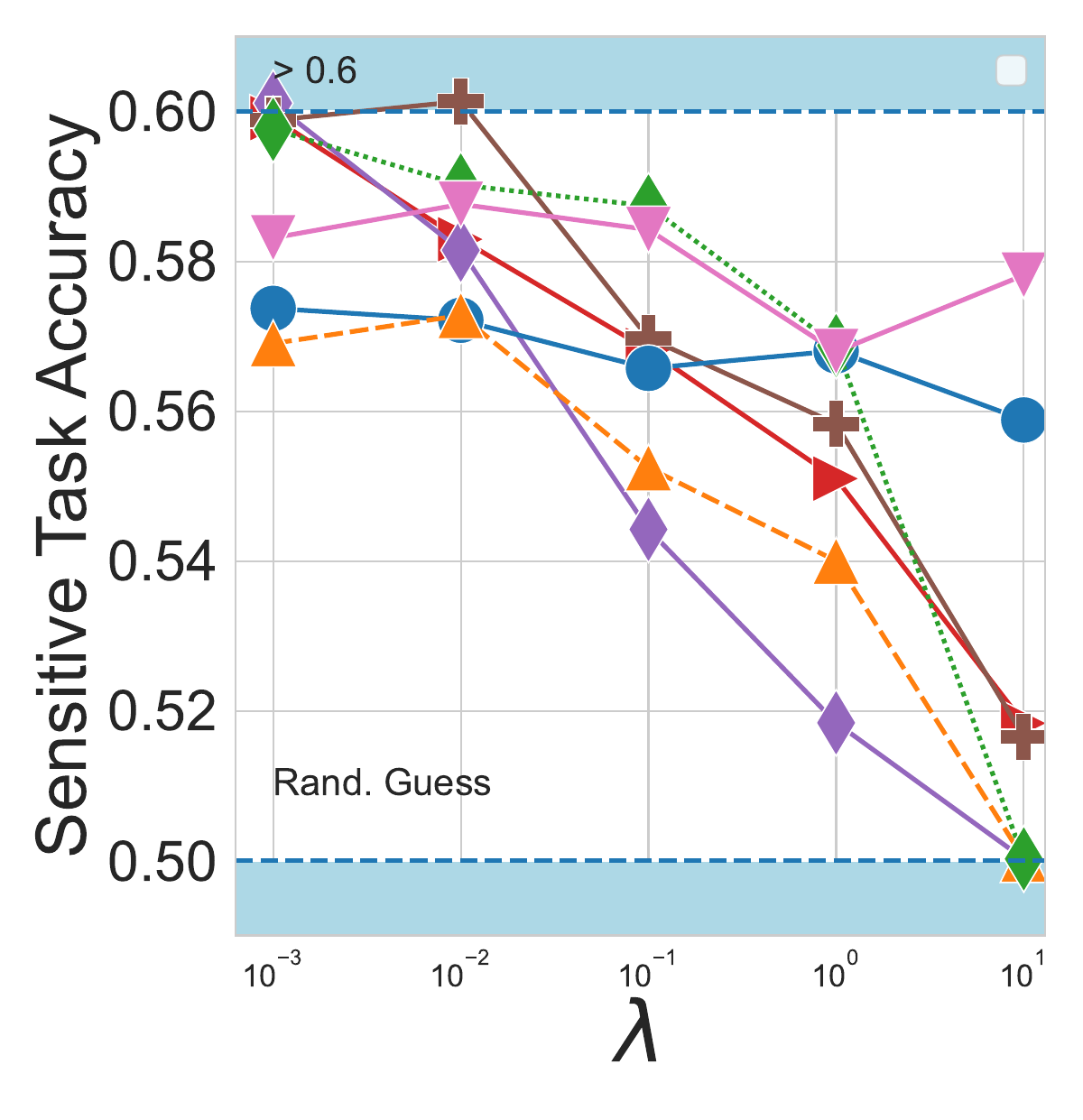}} \\
    \subfloat[\centering $Y$ RNN \texttt{PAN}-Gender ]{{\includegraphics[width=4cm]{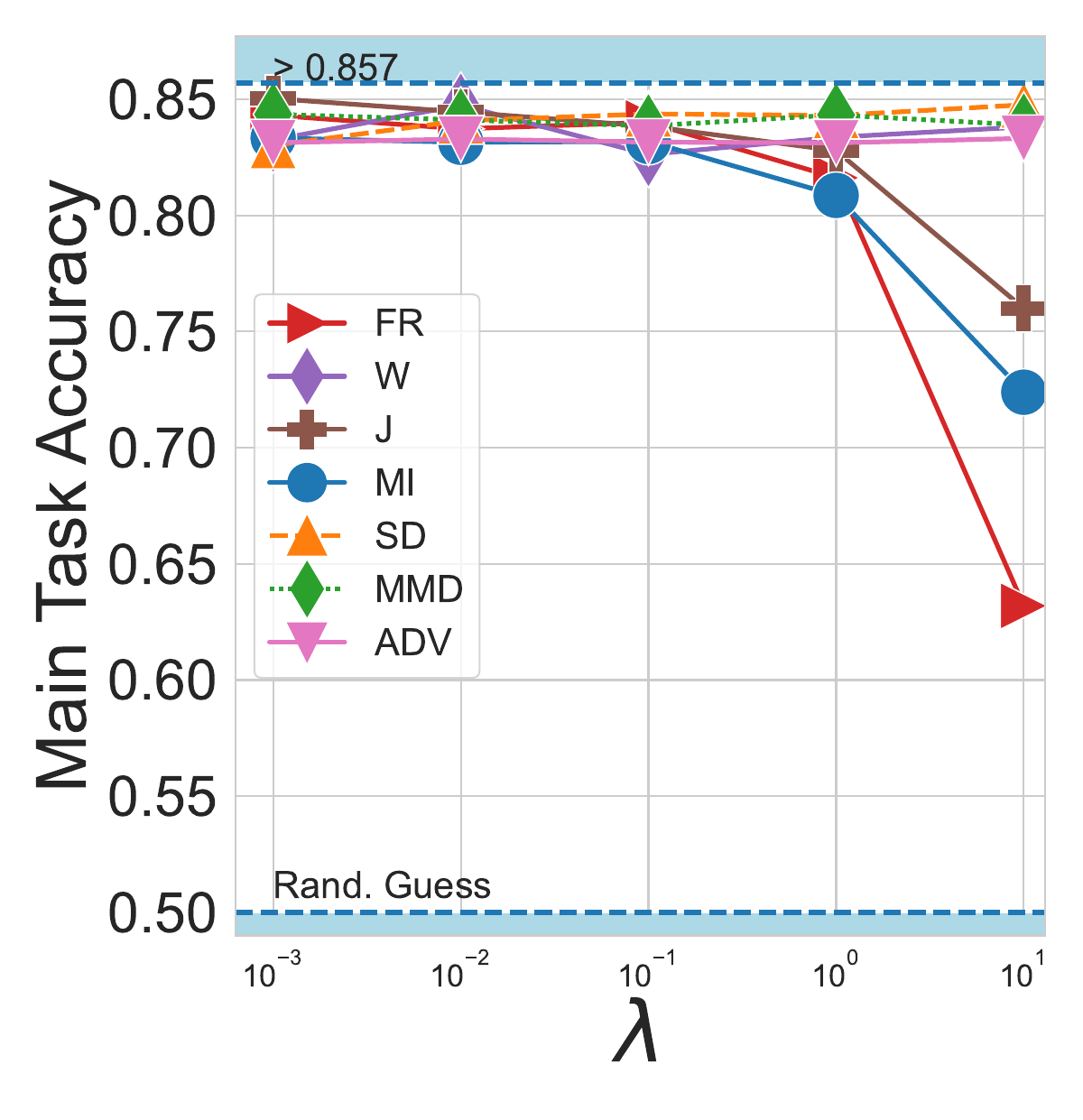} }}%
    \subfloat[\centering $S$ RNN \texttt{DIAL}-Gender ]{{\includegraphics[width=4cm]{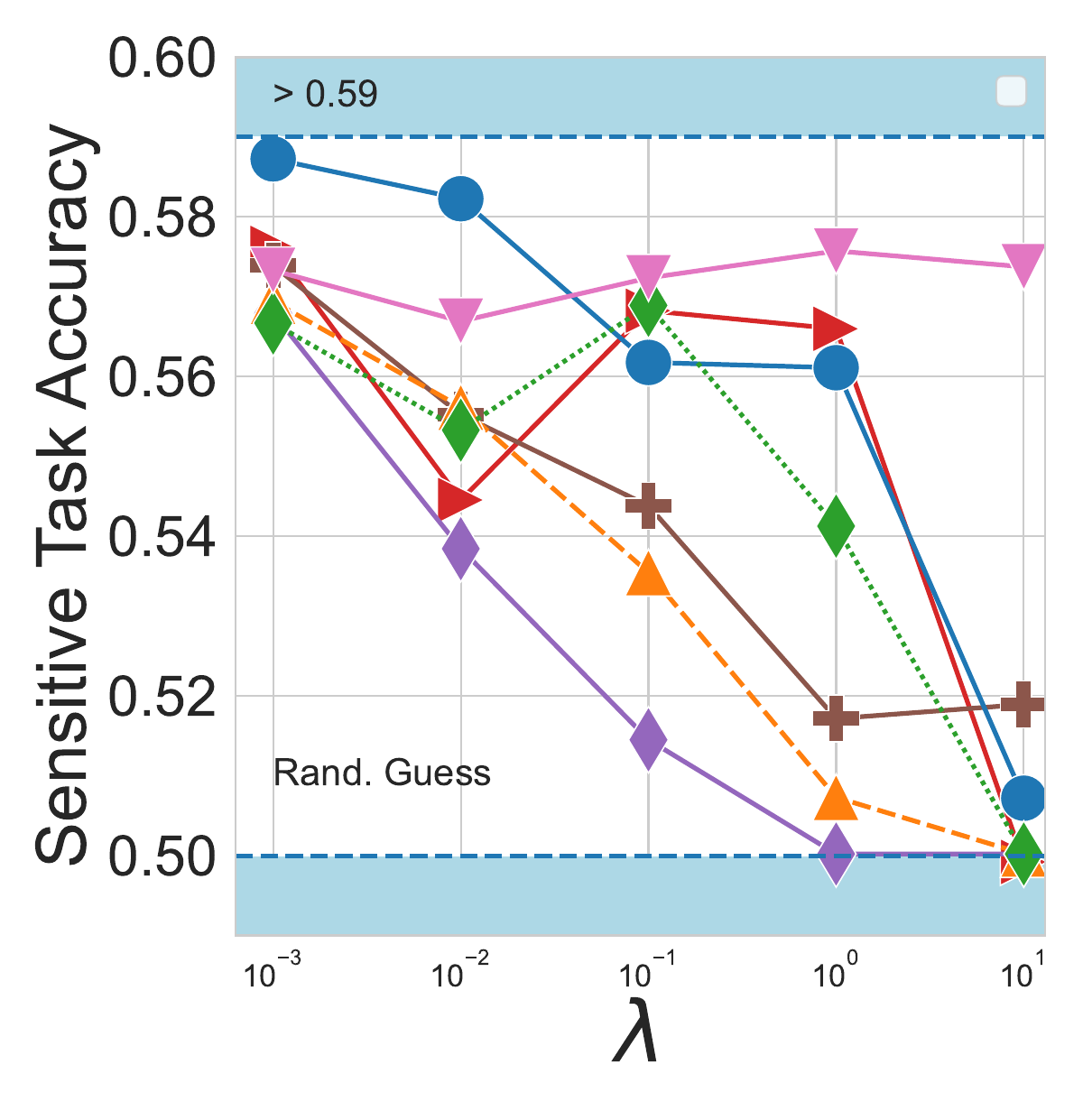} }}%
  \caption{Results on \texttt{PAN} with randomly initialized RNN encoders. Dash lines correspond to model trained with CE loss solely (i.e case $\lambda=0$ in (\ref{eq:all_loss})). Figures on the left are dedicated to the age
attribute while the one on the rights reports results on the
gender attribute. The main task consists in predicting $Y$
thus higher is better. The sensitive task accuracy is obtained
by training a classifier to $S$ on the final representation thus
an ideal model would reach 50\% of accuracy}%
    \label{fig:rnn_pan}%
\end{figure}

\begin{figure}%
    \centering
    \subfloat[\centering $Y$ BERT \texttt{PAN}-Age]{{\includegraphics[width=4cm]{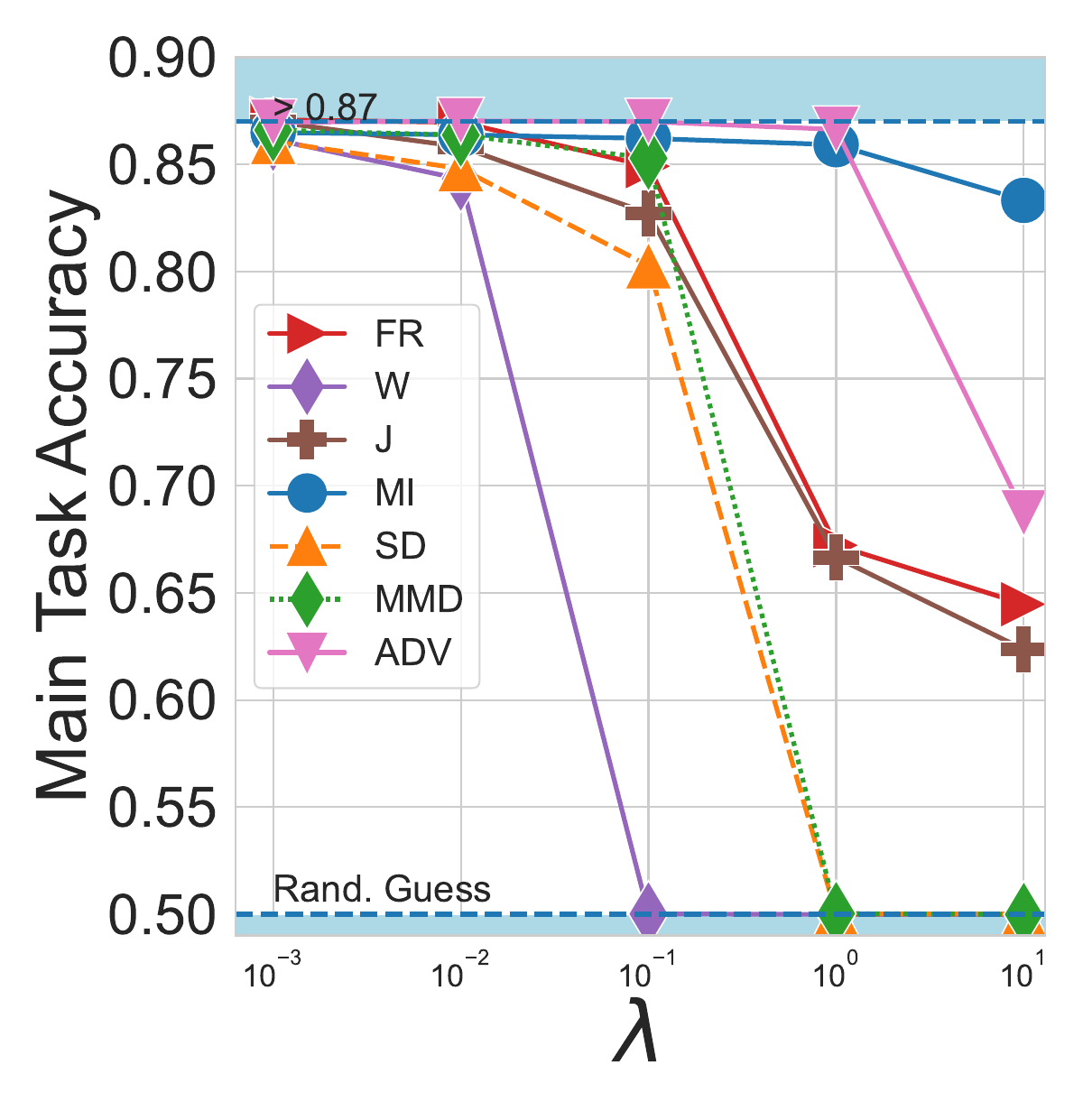} }}%
    \subfloat[\centering $S$ BERT \texttt{PAN}-Age ]{{\includegraphics[width=4cm]{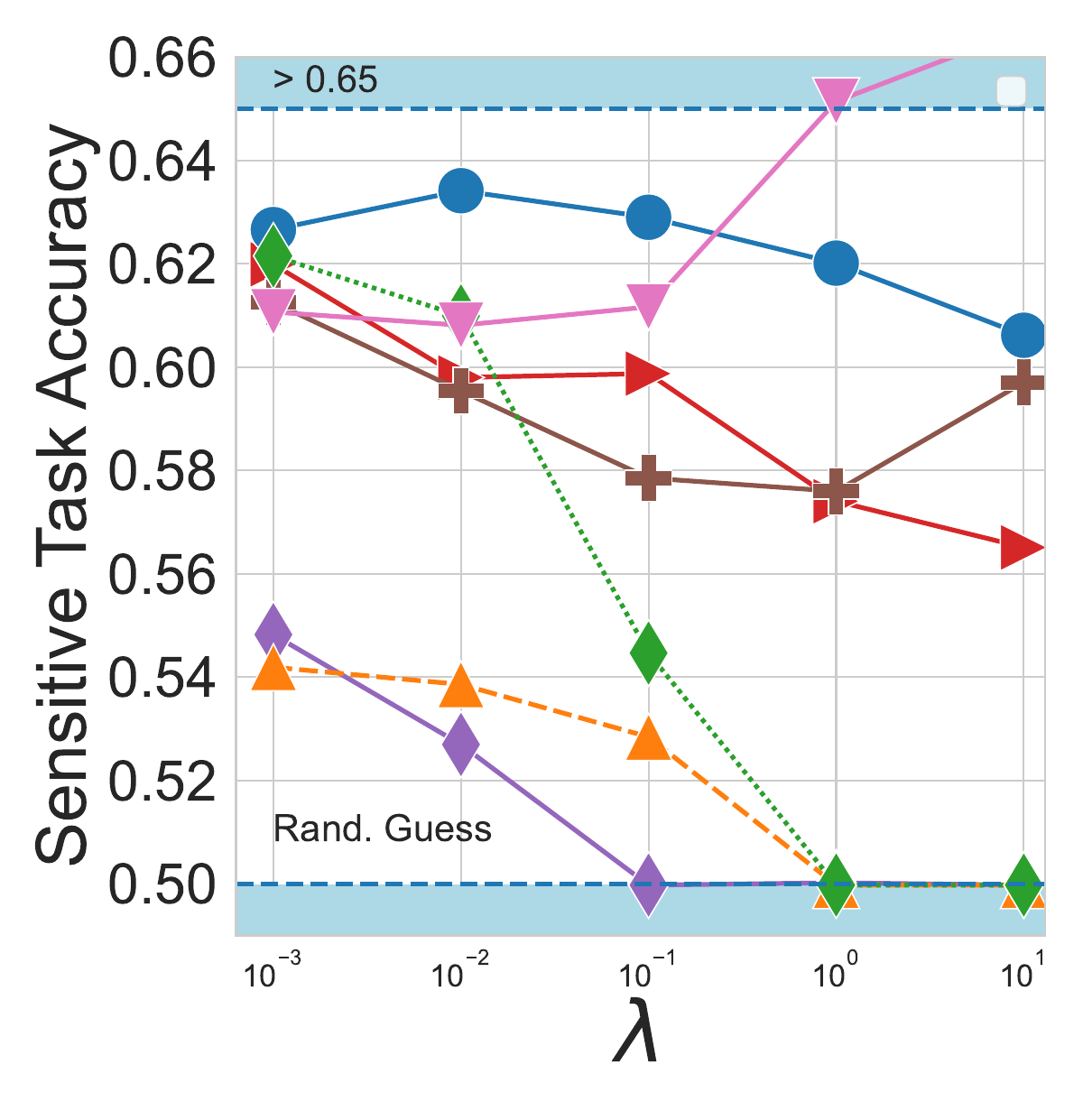} }}\\
    \subfloat[\centering $Y$ BERT \texttt{PAN}-Gender ]{{\includegraphics[width=4cm]{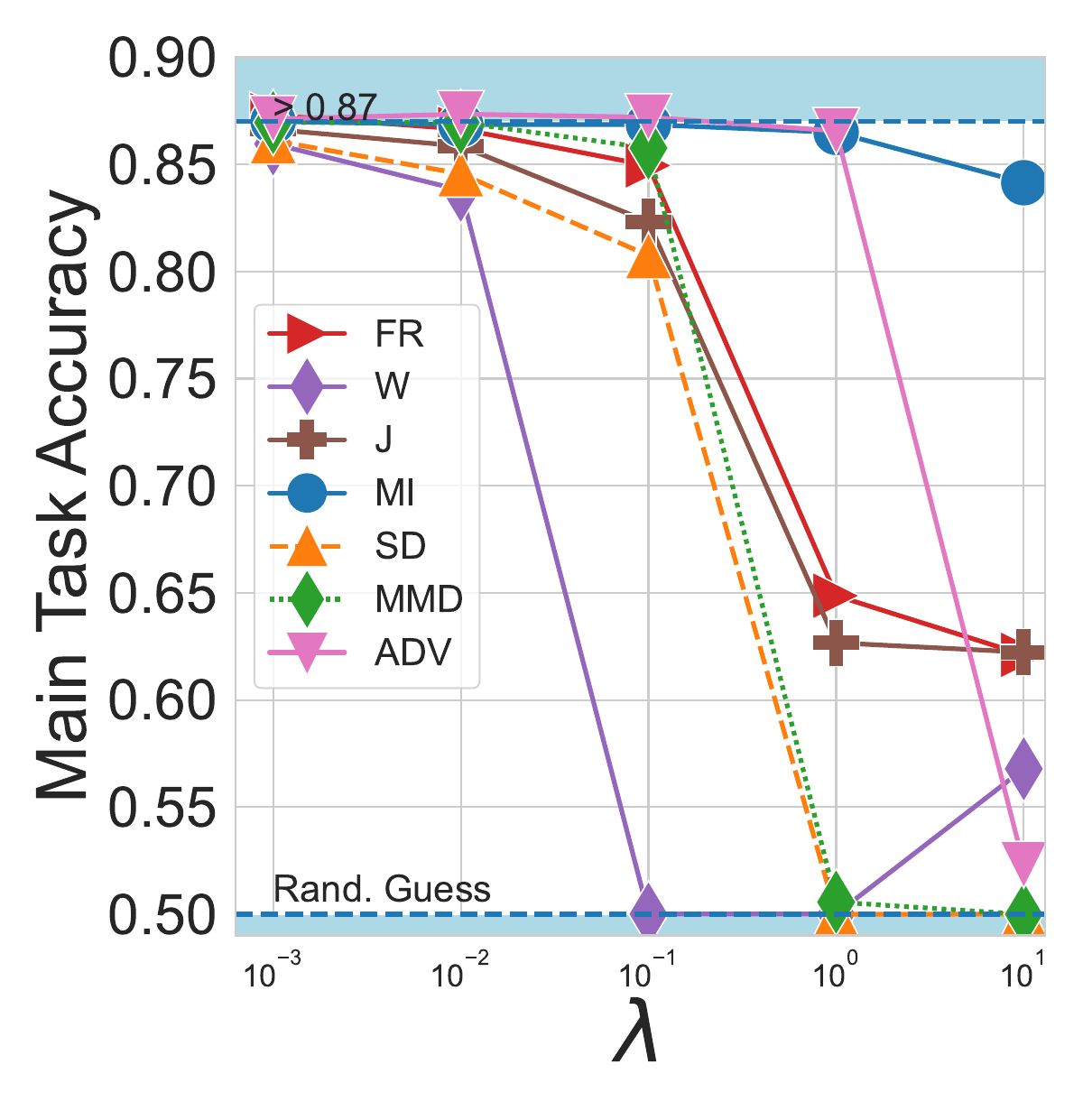} }}%
    \subfloat[\centering $S$ BERT \texttt{PAN}-Gender ]{{\includegraphics[width=4cm]{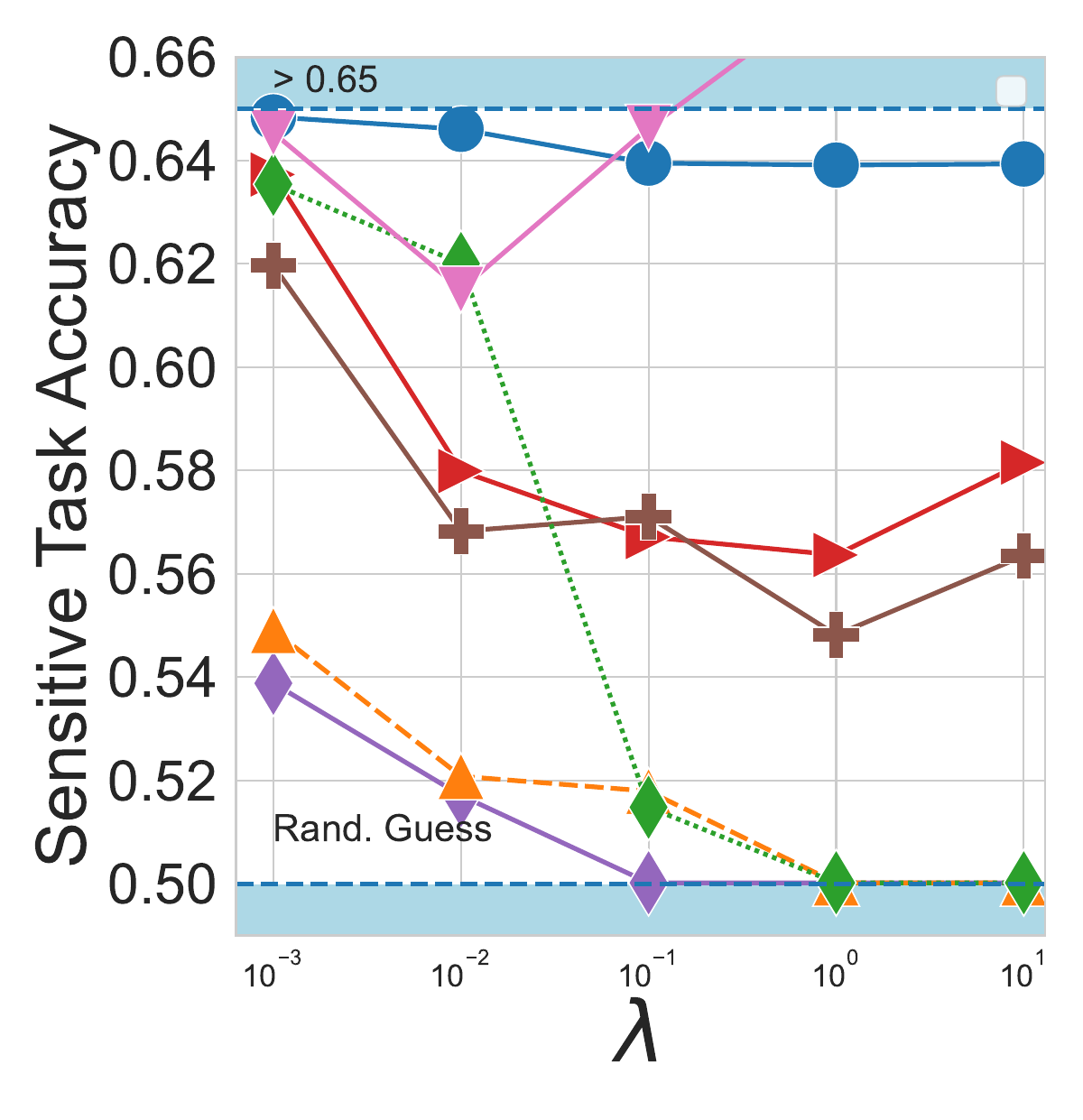} }}%
       \caption{Results on \texttt{PAN} for age (on the left) and gender (on the right) attribute using a pretrained BERT encoder. }%
    \label{fig:bert_pan}%
\end{figure}

 We report in Fig.~\ref{fig:rnn_pan} and Fig.~\ref{fig:bert_pan} the results of the disentanglement analysis on the \texttt{PAN} dataset. 
\\\noindent\textbf{RNN encoders.} We can make the same observations that the one done  on \texttt{DIAL} in Section ~\ref{ssec:overal_results}. We observe that the {W} regularizer performs well and is among the most controllable loss. It is worth noting the good performance of the {SD} and {MMD} losses which both work well on the RNN encoder. 
\\\noindent\textbf{BERT.} For BERT encoder, we observe a similar steep transition than in Section ~\ref{ssec:overal_results} and we can draw similar conclusions. {FR} , W and J fail to disentangle BERT representation with little cost on the main task. {SD} and {MMD}  achieve good results. 
\\\noindent\textbf{Takeaways.} When working with randomly initialized RNN encoders to learn disentangled representation we advise relying on W and when working with pretrained encoder we advise to rely on the {SD}.

\subsection{On the Diagonal Gaussian Assumption}
\begin{figure*}%
    \centering
    \subfloat[\centering RNN - Mention - $S=1$ RNN]{{\includegraphics[width=4cm]{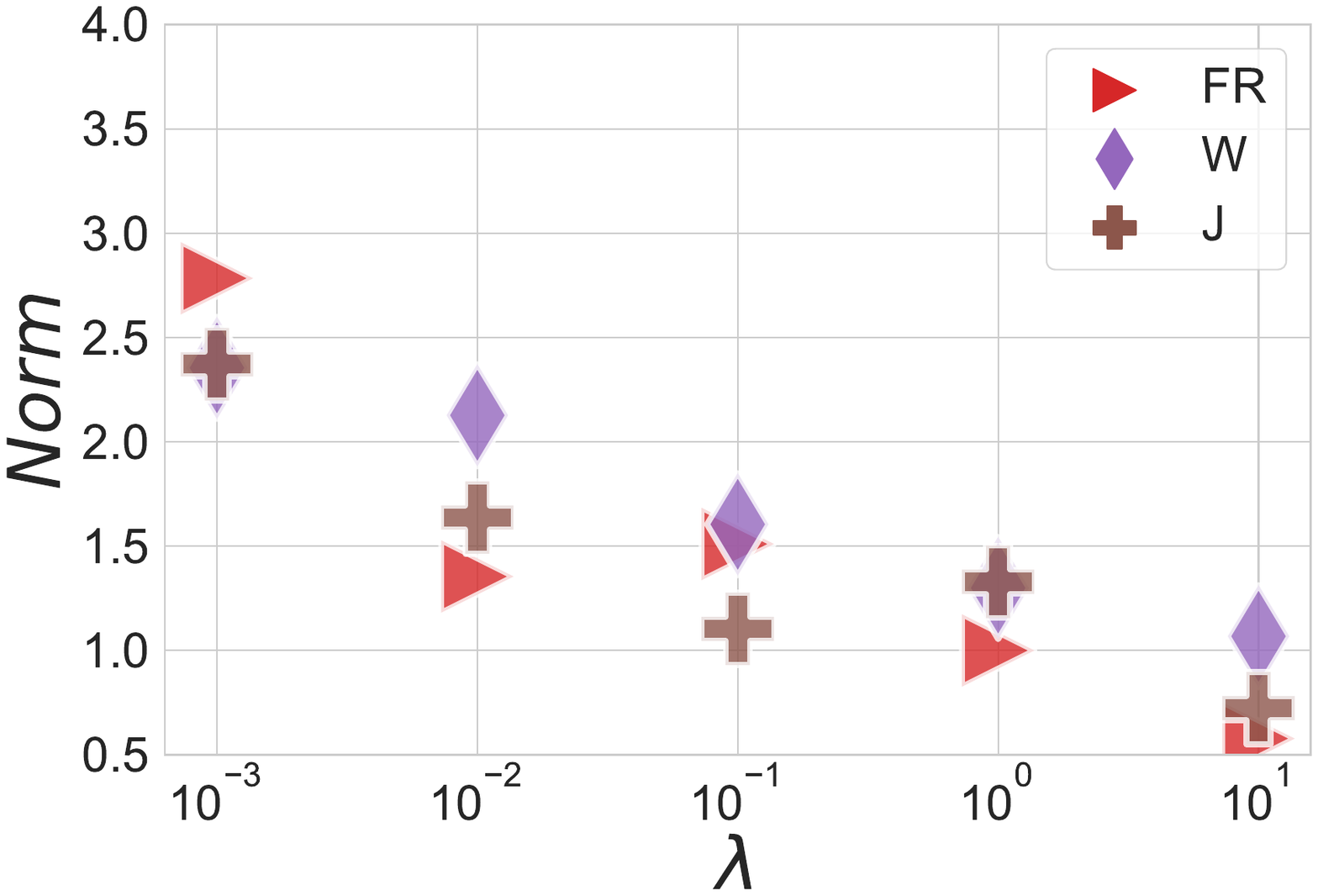}}}%
    \subfloat[\centering RNN - Mention - $S=0$ ]{{\includegraphics[width=4cm]{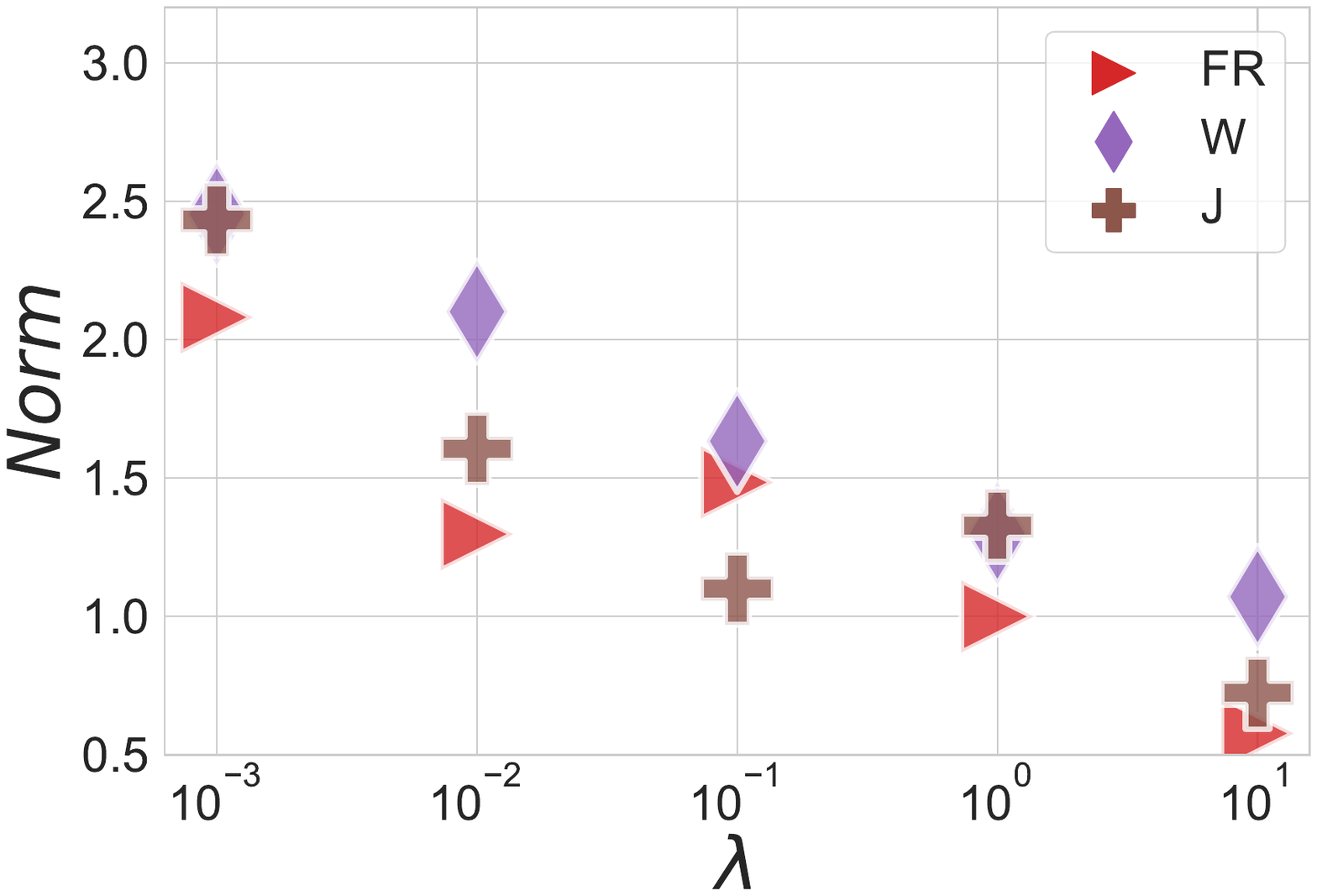}}}%
        \subfloat[\centering BERT - Mention - $S=1$ ]{{\includegraphics[width=4cm]{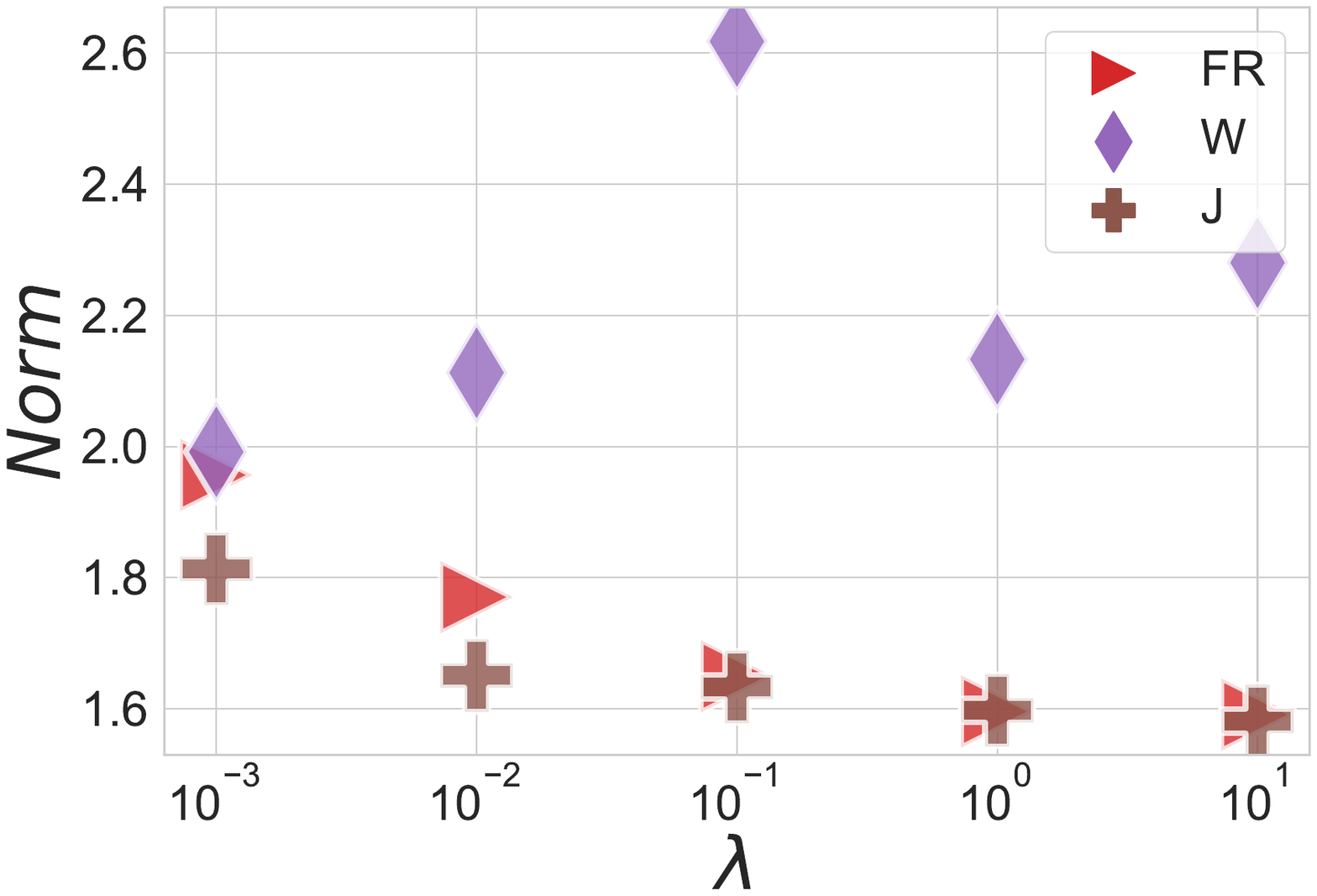}}}%
            \subfloat[\centering BERT - Mention - $S=0$ ]{{\includegraphics[width=4cm]{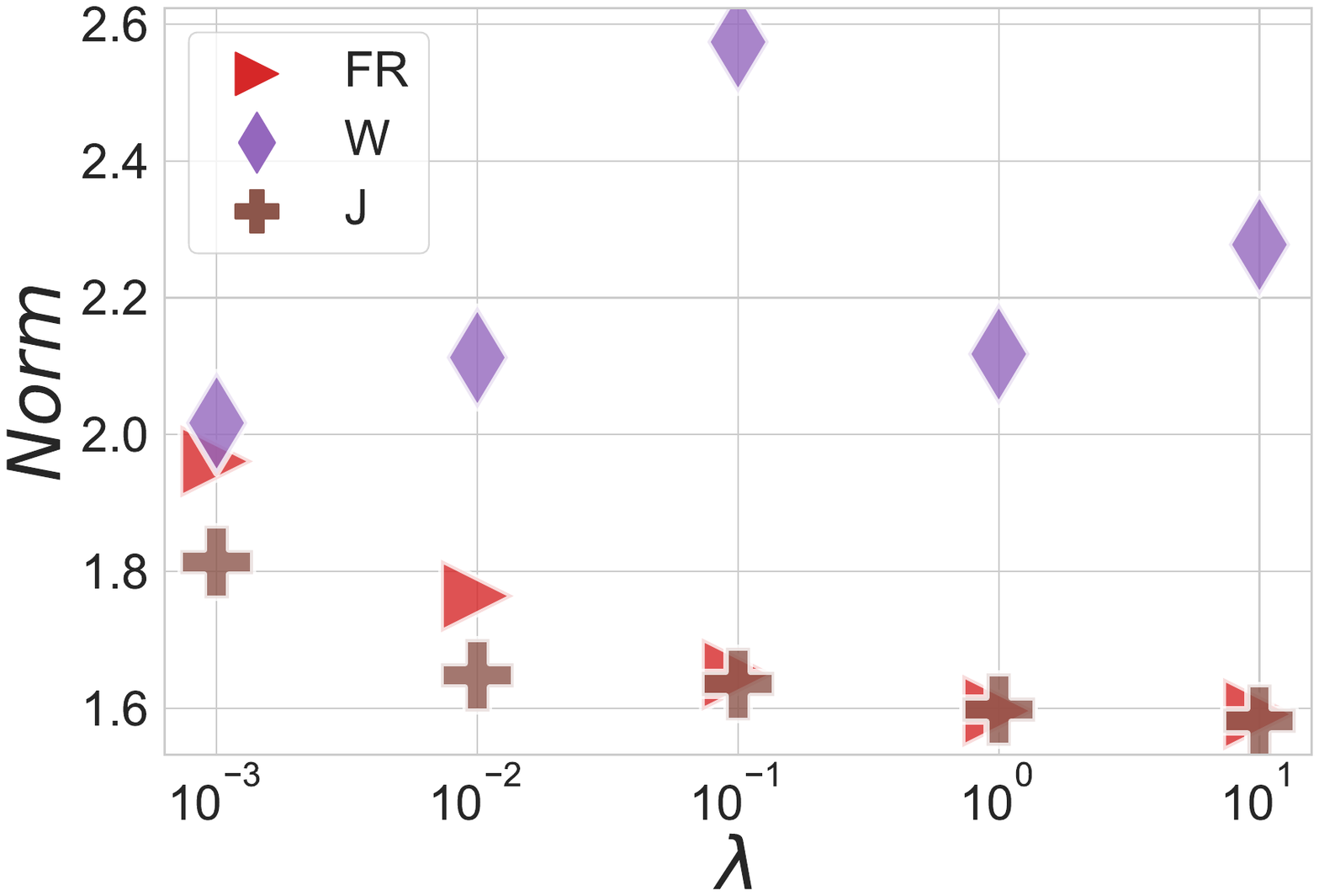}}} \\
    \subfloat[\centering  RNN - Sentiment - $S=1$]{{\includegraphics[width=4cm]{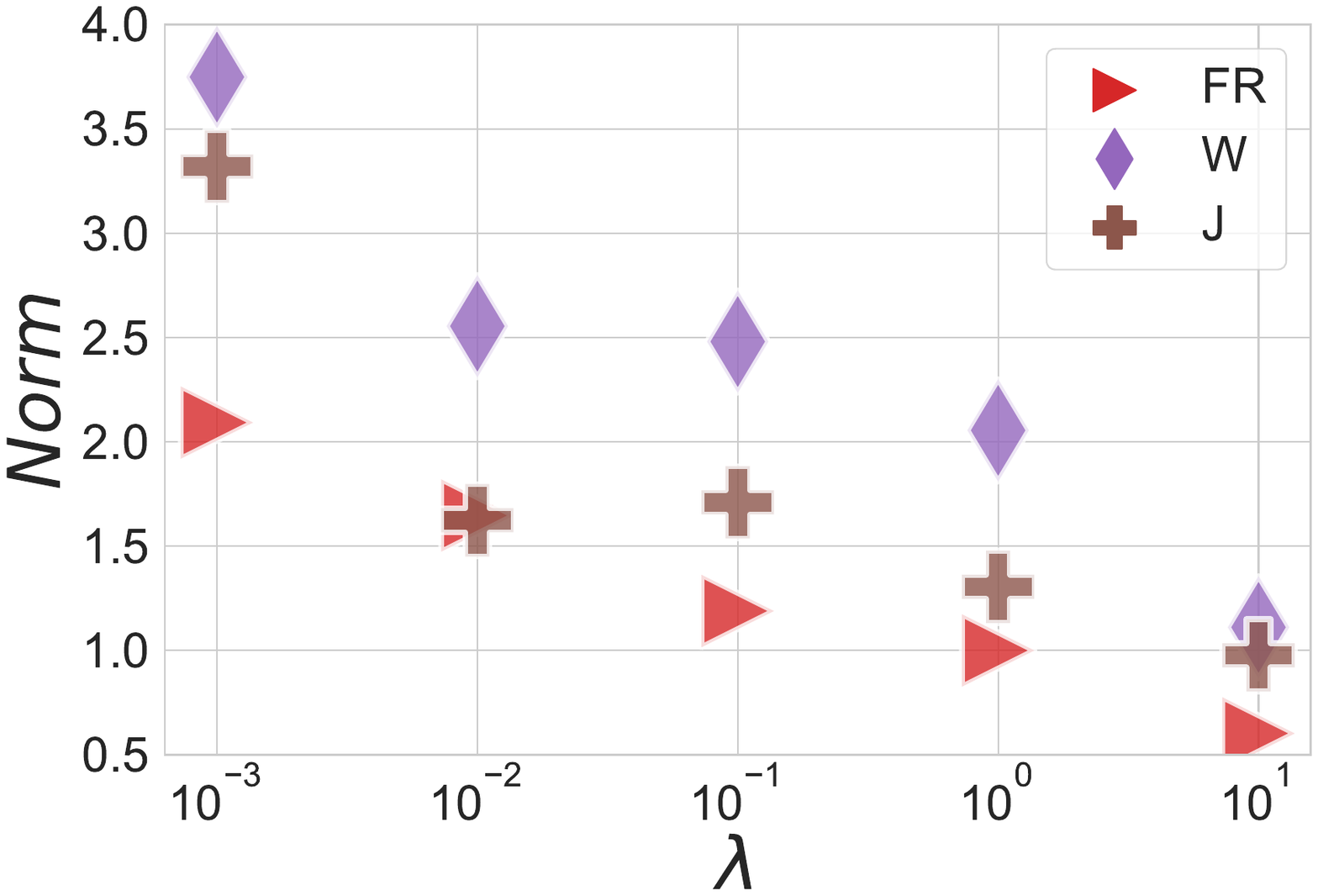}}}%
    \subfloat[\centering  RNN - Sentiment - $S=0$]{{\includegraphics[width=4cm]{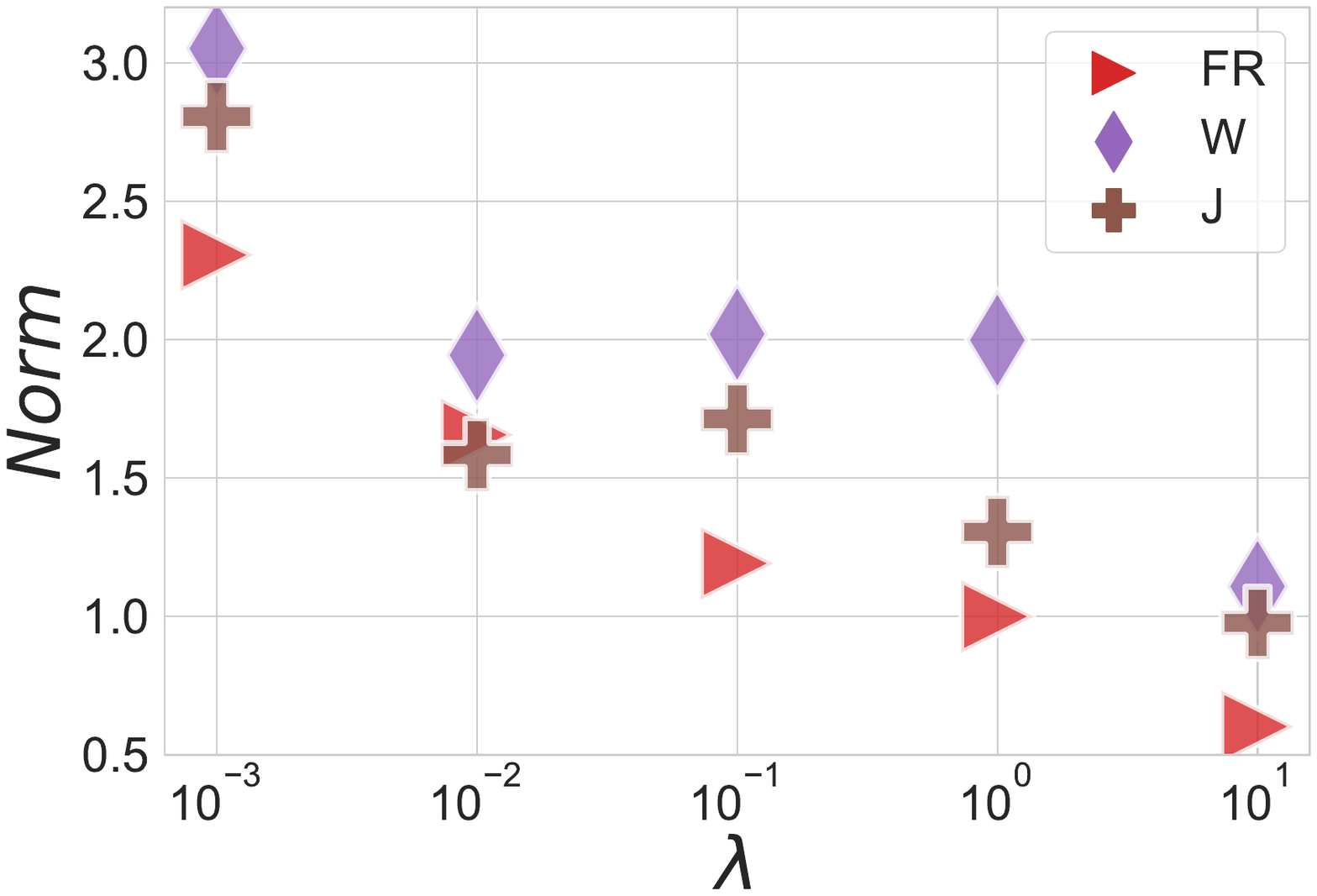}}}%
        \subfloat[\centering BERT - Sentiment - $S=1$ ]{{\includegraphics[width=4cm]{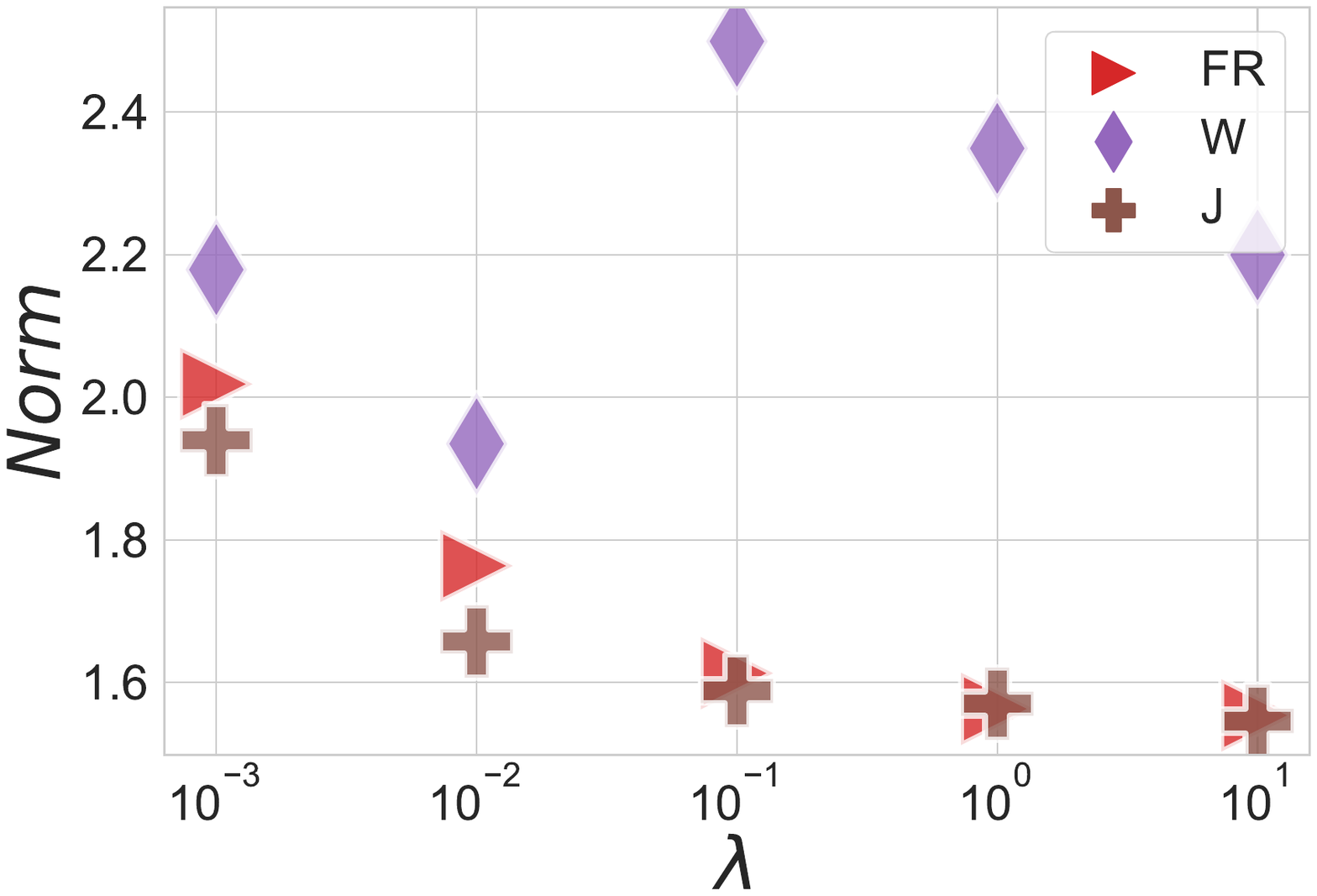}}}%
            \subfloat[\centering BERT - Sentiment - $S=0$ ]{{\includegraphics[width=4cm]{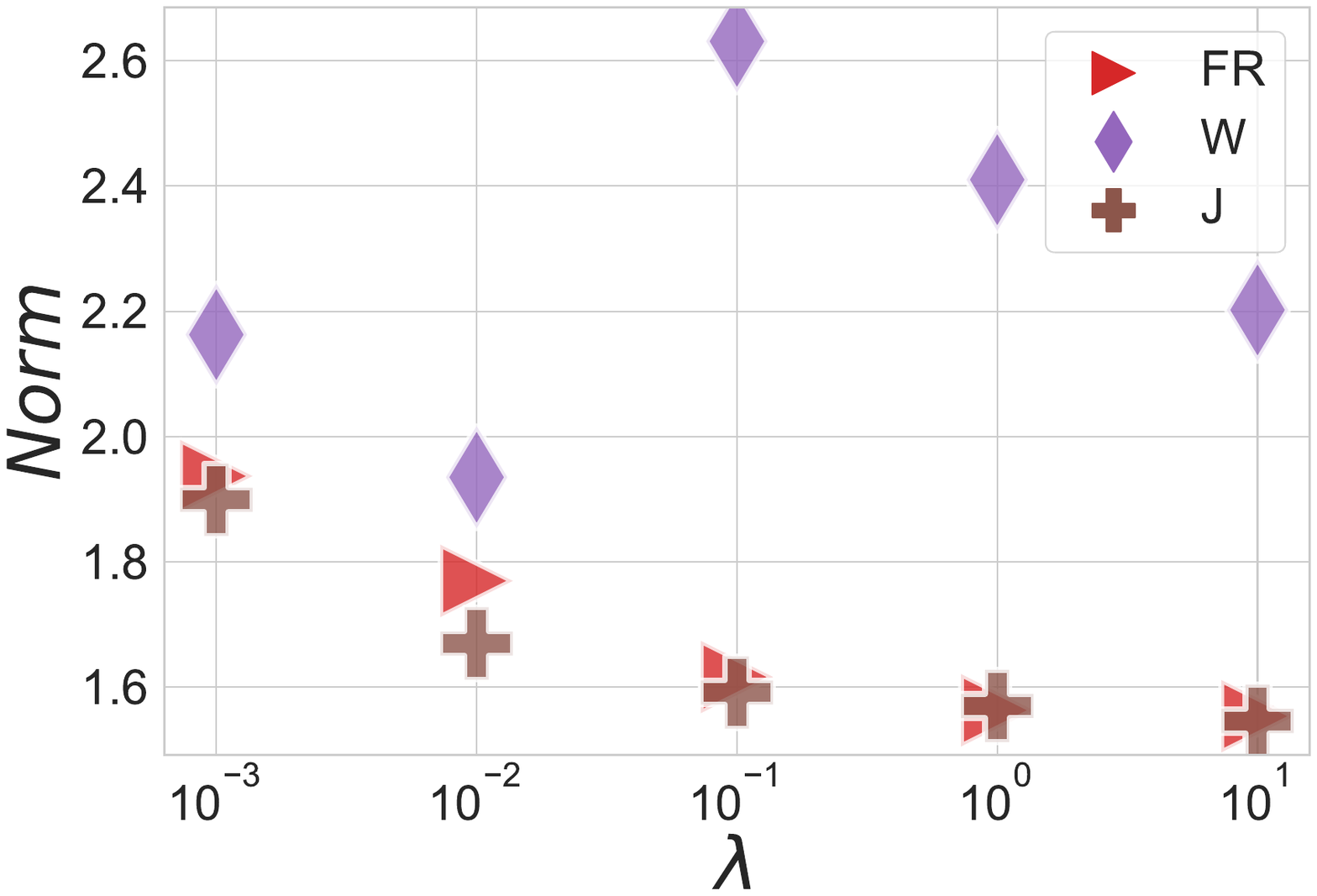}}}%
    \caption{Relative distance as measured by a $L_2$ norm between the empirical covariance matrix and a diagonal matrix on \texttt{PAN} dataset for various values of $S$.}%
    \label{fig:diagonal_all}%
\end{figure*}

Our closed-form for the Fisher-Rao metric relies on the diagonal Gaussian assumption that we have also made for {W} and {J} for a fair comparison. In this experiment (see Fig.~\ref{fig:diagonal_all}), we examine this assumption by evaluating  the relative distance (using a $L_2$-norm) between the empirical covariance matrix and a diagonal matrix.
\\\textbf{Takeaways.} Interestingly, as $\lambda$ increases, the empirical covariance matrix becomes closer to a diagonal matrix. For BERT, we observe that the {W} saturates and the distance for $\lambda > 0.1$ is higher than for RNN. This might be the result of the optimization problems identified in Fig.~\ref{fig:training_dynamic}. Hence, we observe that our methods--when learning more disentangled representations--is that the covariance matrix becomes closer to a diagonal matrix. 


\section{Experimental Details}\label{sec:supp_experimental_details}

\subsection{Replication}
In this section we gather the model details we used in our experiments. All models rely on the tokenizer based on Word Piece \cite{schuster2012japanese} and is similar to the one used for BERT (\textit{i.e} \texttt{bert-base-uncased}) and possess over $40k$ tokens.

\paragraph{Model Architecture for the RNN encoder.} For the randomly initialized RNN encoder, we use a bidirectionnal GRU \cite{gru}  that is composed of 2 layers with an hidden dimension of 128. For activation, we use LeakyReLU \cite{leaky_relu}  and the classification head is composed of fully connected layers of input dimension 256. The learning rate of  AdamW \cite{loshchilov2017decoupled} is set to $0.0001$ and the dropout \cite{dropout}  is set to 0.2. The number of warmup steps \cite{vaswani2017attention}  is set to 1000. 
\paragraph{Computational Resources.} For all 140 models, we train on NVIDIA-V100 with 32GB of RAM. Each model is trained for 30k steps and the model with the best disentanglement accuracy is selected based on the validation set. Each model takes around 5 hours to train. Evaluation requires to train and adversary composed of 3 hidden layers of input 128-128-128-2. The evaluation which involves the training of the probing classifier takes below 1 hour of GPU time. Overall, we train 6 different classifiers per model which correspond to 840 models.

\paragraph{Model Architecture for the BERT encoder.} For the BERT encoder, we add a classification head composed of one fully connected layer. We use a learning rate of $0.00001$ for AdamW and he number of warmup steps is set to 1000. 
\paragraph{Computational Resources.} For all the 140 models, we train on NVIDIA-V100 with 32GB of RAM. Each model is trained for 10k steps, which correspond to the convergence of the model and the model with the best disentanglement accuracy is selected based on the validation set. Each model takes approximately 3 hours to train. Evaluation requires to train and adversary composed of 3 hidden layers and involes LeakyRely and dropout rate of 0.1 of input 768-768-768-2. The evaluation which involves the training of the probing classifier takes below 1 hour of GPU time. Overall, we train 6 different classifiers per model which correspond to 840 models.

\subsection{Negative Results}
We briefly describe a few ideas that did not look promising in our experiments to help future research. Specifically,
\begin{itemize}
\item We attempt to combine our work with MINE from \citet{mine} and we observe high instability during the training.  
    \item We additionally used the clozed-form of {MMD}  under a multivariate Gaussian assumption which lead to poor results (there was no protection against the classifier). 
    \item We also used the Hausdoff distance \cite{serra1998hausdorff} which interpolates between the Iterative closest point \cite{chetverikov2002trimmed} loss and a kernel distance as well as {MMD}  with Laplacian kernel \cite{kondor2016multiscale}. For both case, we ended with optimization issues and poor trade-offs.
\end{itemize}

\subsection{Dataset Examples}
For completness, we gather in this section examples of the \texttt{DIAL} and \texttt{PAN} corpus. Note that this samples have been randomly selected. We report in Table ~\ref{tab:dial_corpus} same randomly sampled examples text from the \texttt{DIAL} corpus and order them based on the sensitive attribute race. The polarity label is obtained through emojis. The goal of the mention task is to predict if a tweet is conversational (i.e., contains a @mentions tokens)
\begin{table*}[]
    \centering
    \begin{tabular}{cc}\hline\hline
    Non-hispanic blacks & Non-hispanic whites \\
    \hline\hline

     \makecell{ain't no beef her and desmond wack dick ass tryna \\ be petty but you know me ion throw slangs}    &  	 
Everyone go get a Vine	 \\
  those r fire red 5s arnt they &    \makecell{ Just exfoliated my face and it 	 \\ feels amazing . \#refreshing \#clean  }  	\\
   \makecell{ Wow that was so deep . I may have teared up a bit . \\ Hahahahah jk that was so fucking gay}   & I've seriously has the worse luck this weekend	\\

  \makecell{ lol Sh*t Get U Where U Need To Go \\ If U In That Situation}	&       \makecell{  Why does my phone take years \\ to update and download apps ?}\\
 Chief Keef - Ain't Done Turning Up ” &   \makecell{ If this tweet gets 1,000 retweets \\ I will get one thousand retweets} \\\hline\hline
    \end{tabular}
    \caption{Randomly sample for \texttt{DIAL} corpus. The sensitive attribute is the race as defined in \citet{dial}.}
    \label{tab:dial_corpus}
\end{table*}

We report in Tab. \ref{tab:pan} examples from the \texttt{PAN} corpus. The age attribute is obtained through birth-date published on the user’s Linkedin profile whereas for the gender the authors rely on both the user’s name and photograph.

\begin{table*}[]
    \centering
   \resizebox{\textwidth}{!}{ \begin{tabular}{cc}\hline\hline
    Above 35 & Below 35 \\
    \hline\hline 
\makecell{It’s amazing ! RT : I need to get to to see \\ this exhibition . Looks brilliant ! \#Photorealism }&
\makecell{Behind the Screens of Twitter's Funniest \\ Parody Accounts http://t.co/siLJo0nkZt} \\

\makecell{So funny when Notting Hill comes on the tele \\ to see and his reaction . \#hisfavfilm \#softoldromantic}&
\makecell{good luck for tomorrow Sean . \#ComeonTheGrugy} \\

\makecell{So long ... hello \#iPhone !}&
\makecell{Super Cheap Papa John's Pizza \\ \#freebies  http://t.co/eMtKikPNnM} \\\hline\hline
    \end{tabular}}
    \caption{Randomly sample from \texttt{PAN} corpus the sensitive attribute is the age. This dataset has been proposed in \citet{rangel2014overview}.}
    \label{tab:pan}
\end{table*}

\subsection{Related Work General Algorithm}\label{subsec:oldalgo}
For completeness we provide in Algorithm ~\ref{alg:fitted_disent2} the algorithm used for training adversarial or MI-based regularizers. It is worth noting that these baselines require extra learnable parameters that need to be tuned using a Nested Loop.

\begin{algorithm}
  \caption{Disentanglement using adversarial-based or MI-based regularizer
 }
  \begin{algorithmic}[1]
   \\\textsc{Input} $\mathcal{D}=\{(x_{j},s_{j},y_{j}), \forall j \in [1,n]\}$,  $\mathcal{D^\prime}=\{(x_{j},s_{j},y_{j}), \forall j \in [n+1,N]\}$, $\theta$
parameters of the encoder network, $\phi$ parameters of the main classifier, $\psi$ parameters of the regularizer.
\\\textsc{Initialize} parameters $\theta$, $\phi$, $\psi$
\\\textsc{Optimization}
\While{$(\theta,\phi,\psi)$ not converged} 
\For{\texttt{ $i \in [1,Unroll]$}} \Comment{Nested loop}
                \State Sample a batch $\mathcal{B^\prime}$ from $\mathcal{D^\prime}$
        \State Update $\phi,\psi$ using (\ref{eq:all_loss}).
      \EndFor
\State Sample a batch $\mathcal{B}$ from $\mathcal{D}$
\State Update $\theta$ with $\mathcal{B}$ (\ref{eq:all_loss}).
\EndWhile
\\\textsc{Output}  Encoder and classifier weights ${\theta,\phi}$
  \end{algorithmic}
  \label{alg:fitted_disent2}
\end{algorithm}

\subsection{Future Work.}
As future work we plan to disentangled more complex labels such as dialog acts \cite{colombo2020guiding,DBLP:conf/emnlp/ColomboCLC21a}, emotions \cite{DBLP:conf/wassa/WitonCMK18} and linguistic phenomena such as disfluencies \cite{DBLP:conf/emnlp/DinkarCLC20} and other spoken language phenomenon \cite{DBLP:conf/emnlp/ChapuisCMLC20}. Future research also include extending these losses to data augmentation \cite{DBLP:journals/corr/abs-2112-02721,DBLP:conf/icnlsp/ColomboCYV21} and sentence generation \cite{DBLP:journals/corr/abs-2112-01589,DBLP:conf/emnlp/ColomboSCP21} and study the trade-off using rankings \cite{DBLP:journals/corr/abs-2202-03799} or anomaly detection \cite{FIF,ACH,staerman2021affine,staerman2022functional}.

\subsection{Libraries used.}
For this project among the library we used we can cite: 
\begin{itemize}
    \item Transformers from \cite{hugging_face}.
    \item Pytorch \cite{pytorch} for the GPU support.
    \item Geomloss \cite{feydy2019interpolating} for the {SD} and {MMD}. It can be found at {\url{https://www.kernel-operations.io/geomloss}}
\end{itemize}

\end{document}